%% file: neurips_2026.tex
\documentclass{article}

 \usepackage[preprint]{neurips_2026}

\usepackage[utf8]{inputenc} 
\usepackage[T1]{fontenc}    
\usepackage{hyperref}       
\usepackage{url}            
\usepackage{booktabs}       
\usepackage{amsfonts}       
\usepackage{nicefrac}       
\usepackage{microtype}      
\usepackage{xcolor}         
\usepackage{graphicx}
\usepackage{caption}
\usepackage{colortbl} 
\usepackage{pifont}
\usepackage{amsmath}
\usepackage[table]{xcolor}
\usepackage[most]{tcolorbox}
\usepackage{bbm}
\usepackage{multirow}
\tcbuselibrary{breakable, skins}
\usepackage{marvosym}

\newcommand{\cmark}{\ding{51}}
\newcommand{\xmark}{\ding{55}}
\newcommand{\pmark}{\textcolor{gray}{\(\sim\)}}
\newcommand{\caseicon}[1]{\raisebox{-1pt}{\includegraphics[height=1em]{#1.pdf}}}

\definecolor{WanBlue}{HTML}{3D7BFF}
\definecolor{WanLight}{HTML}{E6EEFF}

\definecolor{mechanics}{HTML}{798FCA}
\definecolor{fluid}{HTML}{AECDB1}
\definecolor{materials}{HTML}{E7BE90}
\definecolor{thermal}{HTML}{E4ABA6}
\definecolor{chemistry}{HTML}{BCDAE1}
\definecolor{electromagnetism}{HTML}{A694D1}

\newcommand{\phymetric}{\textsc{PhyMetric}}

\newtcolorbox{categorycardbox}[2][]{%
  enhanced, breakable, before skip=6pt, after skip=6pt,
  colback=#2!4, colframe=#2!70!black,
  boxrule=0.6pt, arc=2pt, left=5pt, right=5pt, top=3pt, bottom=3pt,
  fonttitle=\footnotesize\bfseries, 
  fontupper=\footnotesize,           
  #1
}
\newcommand{\catfield}[2]{\textbf{#1:}~{#2}\par}

\newcommand{\categorycard}[9]{%
  \begin{categorycardbox}{#3}[title={#1\hfill\normalfont\textit{Domain:} #2 \quad \textit{Scenes:} #9}]\\
    \catfield{Description}{#4}
    \catfield{Physical principle}{#5}
    \catfield{Object pool}{\textit{#6}}
    \catfield{Environment pool}{\textit{#7}}
    \catfield{Key dimensions}{#8}
  \end{categorycardbox}
}

\newtcolorbox{evalcasebox}[1][]{%
  enhanced, breakable, before skip=8pt, after skip=8pt,
  colback=green!1, colframe=green!48!black!50,
  boxrule=0.6pt, arc=2pt, left=5pt, right=5pt, top=3pt, bottom=3pt,
  fonttitle=\bfseries, #1
}

\newcommand{\evalframerow}[6]{%
  \begin{center}
  \begin{tabular}{@{}c@{\hspace{4pt}}c@{\hspace{4pt}}c@{}}
    \includegraphics[width=0.30\linewidth, height=0.18\linewidth, keepaspectratio]{#1} &
    \includegraphics[width=0.30\linewidth, height=0.18\linewidth, keepaspectratio]{#3} &
    \includegraphics[width=0.30\linewidth, height=0.18\linewidth, keepaspectratio]{#5} \\[1pt]
    {\scriptsize #2} & {\scriptsize #4} & {\scriptsize #6} \\
  \end{tabular}
  \end{center}
}

\title{PhyMetric: Diagnosing Physical Plausibility in Text-to-Video Generation via Scene-Level QA}

\author{%
  Siwei Meng \qquad
  Xuerou Cai \qquad
  Rui Hu \qquad
  Ping Liu\textsuperscript{\Letter}  \\
  Department of Computer Science \& Engineering \\
  University of Nevada, Reno \\
  Reno, NV 89512 \\
}

\begin{document}

\maketitle

\begingroup
\renewcommand{\thefootnote}{\Letter}
\footnotetext{Corresponding author: Ping Liu
(\texttt{pino.pingliu@gmail.com}).}
\endgroup
\begin{abstract}
Text-to-video (T2V) generation models can produce visually realistic videos that nevertheless violate the physical consequences implied by the prompt and scene.
Existing benchmarks largely rely on holistic plausibility scores or broad quality dimensions, leaving it unclear which physical behavior is being tested and why a model fails.
We introduce \phymetric{}, a scene-level diagnostic benchmark that recasts physical plausibility evaluation as verifiable, scene-grounded question answering.
\phymetric{} contains 1{,}115 scenes spanning 50 physical categories across six domains, paired with 8{,}920 answer-balanced binary questions over 12 evaluation dimensions.
Each scene is associated with eight targeted yes/no questions and a structured physical prior specifying the governing law, expected temporal sequence, observable signatures, and common failure modes.
At evaluation time, a vision-language model (VLM) judge answers each question from question-aware sampled frames conditioned on the prior, and correctness is hierarchically aggregated into scene, dimension, physical category, and model scores.
Experiments on four open-source T2V models and three VLM judges show that scene-specific binary QA is a critical component of the protocol: when each targeted question is replaced by a single generic plausibility query, evaluator accuracy drops to the chance level.
Structured physical priors offer a complementary signal that improves human alignment of the resulting scores, with effects that vary across VLM judges of different capacity.
These findings establish \phymetric{} as a benchmark that turns physical plausibility evaluation from opaque holistic scoring into interpretable, scene-level inspection. 
Our project page is available at: \url{https://mengsiwei.github.io/PhyMetric/}.

\end{abstract}

\input{sections/01introduction}
\input{sections/02relatedWorks}

\input{sections/03benchmark}

\input{sections/05experiment}

\input{sections/07conclusion}

\clearpage

\bibliographystyle{unsrt} 
\bibliography{reference}
\clearpage

\appendix

\section{Technical appendices and supplementary material}

In this appendix, we provide additional technical details, prompts, and experimental results to support the main manuscript. The content is organized as follows:

\begin{description}
    \item[\ref{appen:taxonomy}\quad Physics Taxonomy and Dimensions] \hfill
    \begin{itemize}
        \item 50 physics categories. 
        \item 12 evaluation dimensions.
    \end{itemize}
    
    \item[\ref{appen:scene_construction}\quad Scene and Question Construction] \hfill
    \begin{itemize}
        \item System and user prompts for data generation.
        \item Detailed per-category construction.
    \end{itemize}

    \item[\ref{appen:priors_construction}\quad Structured Physical Priors Construction] \hfill
    \begin{itemize}
        \item Evaluation of various LLM backbones.
        \item The system prompts used to generate physics-grounded priors.
    \end{itemize}

    \item[\ref{appen:vlm_eval}\quad \phymetric{} Evaluation] \hfill
    \begin{itemize}
        \item Prompting strategies for VLM judges.
        \item Technical specifications of the VLM judges and baseline video generators.
        \item Cases of \phymetric{} evaluation.
    \end{itemize}

    \item[\ref{appen:exp_details}\quad Experimental Details] \hfill
    \begin{itemize}
        \item Ablation study setups and question query formulation.
        \item Methodology and participant demographics for the human study.
    \end{itemize}

    \item[\ref{appen:limitation}\quad Limitations and Future Work] \hfill
    \begin{itemize}
        \item Limitations.
        \item Future work.
    \end{itemize}
\end{description}
\newpage
\subsection{Physics Taxonomy and Dimensions}
\label{appen:taxonomy}

\paragraph{Physics domains.}
\phymetric{} groups its 50 categories into six physics domains so that aggregate analysis can isolate where T2V models succeed or fail at the level of broad physical theory rather than individual phenomena.
Table~\ref{tab:domain_breakdown} summarizes the per-domain category and scene counts; the full per-category card listing is in \S\ref{app:category_cards}.

\begin{table}[h]
\centering
\small
\setlength{\tabcolsep}{6pt}
\caption{Six physics domains in \phymetric{} with category and scene counts.}
\label{tab:domain_breakdown}
\begin{tabular}{lccl}
\toprule
\textbf{Domain} & \textbf{Categories} & \textbf{Scenes} & \textbf{Representative phenomena} \\
\midrule
Mechanics                 & 12 & 280 & gravity, collisions, friction, projectile, pendulum \\
Fluid Dynamics            & 10 & 230 & buoyancy, viscosity, splash, vortex, granular flow \\
Material                  &  8 & 180 & fracture, slicing, cloth, smoke, explosion \\
Thermal                   &  8 & 165 & melting, boiling, freezing, convection, condensation \\
Chemistry                 &  7 & 145 & diffusion, adsorption, dissolution, precipitation \\
Electromagnetism          &  5 & 115 & reflection, refraction, shadow, electrostatics, magnetism \\
\midrule
\textbf{Total}            & \textbf{50} & \textbf{1{,}115} & --- \\
\bottomrule
\end{tabular}
\end{table}

\paragraph{Question dimensions.}
Every binary question is tagged with one or two of the 12 \emph{question dimensions} listed in Table~\ref{tab:dimensions}.
Dimensions are not categories: a single category (e.g., \emph{Splash}) generates questions across several dimensions (\emph{motion}, \emph{shape}, \emph{spatial}, \emph{causal}), and a single dimension (e.g., \emph{motion}) is exercised in dozens of categories.

\begin{table}[h]
\centering
\small
\setlength{\tabcolsep}{6pt}
\caption{The 12 question dimensions used to tag every binary question. Each question carries 1--2 dimension tags; each scene covers $\geq$3 distinct dimensions across its 8 questions.}
\label{tab:dimensions}
\begin{tabular}{lp{0.78\linewidth}}
\toprule
\textbf{Dimension} & \textbf{What it asks about} \\
\midrule
existence            & whether a specific object, region, or event is present in the video \\
size                 & absolute or relative size of an object or region \\
appearance/texture   & color, surface pattern, transparency, sheen, smoothness \\
shape                & geometric form and how it changes during the event \\
motion               & whether/how something moves, including speed and acceleration \\
number               & object count, including stages or instances of a repeating event \\
direction            & axis or sense of motion (up/down, left/right, inward/outward) \\
material             & physical material identity inferred from visual cues \\
object               & object identity and persistence across frames \\
spatial              & relative position, contact, containment, layering \\
interaction          & contact-driven outcomes (collision, push, cut, attract) \\
causal               & whether one event correctly produces or follows another \\
\bottomrule
\end{tabular}
\end{table}

\subsection{Scene and Question Construction}
\label{appen:scene_construction}

\subsubsection{System and user prompt}

\begin{tcolorbox}[
  enhanced, breakable, colback=gray!5, colframe=black!55,
  title=\textbf{System Prompt \texttt{(SCENE\_QUESTION\_GENERATION)}},
  fonttitle=\bfseries, left=4pt, right=4pt, top=2pt, bottom=2pt,
  boxrule=0.6pt, arc=2pt
]
\small\ttfamily
You are an expert physicist and video benchmark designer. Generate physically accurate, visually diverse scene descriptions for text-to-video model evaluation. Respond ONLY with valid JSON.
\end{tcolorbox}

\begin{tcolorbox}[
  enhanced, breakable, colback=blue!3, colframe=blue!88!black!55,
  title=\textbf{User Prompt Template \texttt{(SCENE\_QUESTION\_GENERATION)}},
  fonttitle=\bfseries, left=4pt, right=4pt, top=2pt, bottom=2pt,
  boxrule=0.6pt, arc=2pt
]
\scriptsize\ttfamily
Physical phenomenon: \{category\_name\} \\
Description: \{description\} \\
Physical principle: \{physical\_principle\} \\
Physics domain: \{physics\_domain\} \\[2pt]
Object pool (choose from or invent similar): \\
\hspace*{1em}\{object\_pool\} \\[2pt]
Environment pool: \\
\hspace*{1em}\{environment\_pool\} \\[2pt]
Already existing scene captions (DO NOT duplicate, ensure diversity): \\
\{existing\_captions\} \\[4pt]
Generate exactly \{n\_scenes\} new, diverse scenes as a JSON array. \\
Each scene object must have these exact keys: \\[2pt]
\{ \\
\hspace*{1em}"scene\_name": "<short 3-7 word title>", \\
\hspace*{1em}"scene\_caption": "<1-2 sentence description for a T2V prompt>", \\
\hspace*{1em}"generation\_prompt": "<2-3 sentence prompt with object, setting, action, expected outcome>", \\
\hspace*{1em}"causal\_chain": \{ \\
\hspace*{2em}"initial\_state": "<what is present before the event>", \\
\hspace*{2em}"process": "<what physical process happens during the clip>", \\
\hspace*{2em}"final\_state": "<what state objects reach by end of the 4-s clip>" \\
\hspace*{1em}\}, \\
\hspace*{1em}"questions": [ \\
\hspace*{2em}\{ "id": "01", \\
\hspace*{3em}"question\_text": "<yes/no question about what should be observed>", \\
\hspace*{3em}"dimensions": ["<one or two of: existence, size, appearance/texture, shape,\\
\hspace*{6em}motion, number, direction, material, object, spatial, interaction, causal>"], \\
\hspace*{3em}"difficulty": "<easy|medium|hard>", \\
\hspace*{3em}"frame\_indices": [<2-5 frame indices from 0-119>], \\
\hspace*{3em}"answer": <1 for yes, 0 for no> \}, \\
\hspace*{2em}... (8 questions total: EXACTLY 4 with answer=1 and EXACTLY 4 with answer=0) \\
\hspace*{1em}] \\
\} \\[4pt]
Rules for questions: \\
\hspace*{1em}1. Question number: The number of questions for each scenario is 8, with a positive and negative sample ratio of approximately 1:1. \\
\hspace*{1em}2. answer=1: observable facts a correctly generated video SHOULD show. \\
\hspace*{1em}3. answer=0: physically INCORRECT behaviors that a bad video might show. \\
\hspace*{1em}4. Vary difficulty: at least 2 easy, 2 medium, 2 hard per scene. \\
\hspace*{1em}5. Frame indices from [0, 119] (4-s clip at 30 fps); use [20, 60, 100] for spread. \\
\hspace*{1em}6. Cover at least 3 different dimensions across the 8 questions. \\
\hspace*{1em}7. Questions must be concrete and visually answerable. \\
\hspace*{1em}8. Do not repeat the same question pattern across all 8 questions. \\[2pt]
Respond with a JSON array of \{n\_scenes\} scene objects only. No markdown, no extra text.
\end{tcolorbox}

\subsubsection{Per-category construction details}
\label{app:category_cards}

Each card below lists the inputs that parameterize the prompt above for one physical category: the natural-language description, the governing physical principle, the object and environment pools sampled for variety, the key diagnostic dimensions emphasized in the questions, and the target number of scenes.

\paragraph{Mechanics.}

\categorycard{Gravity\_Earth}{Mechanics}{mechanics}
{Object released and falls under Earth's gravitational acceleration ($9.8\,\text{m}/\text{s}^2$).}
{Newton's second law with constant downward acceleration $g=9.8\,\text{m}/\text{s}^2$; velocity grows linearly, distance as $\tfrac{1}{2}gt^2$.}
{stone, apple, metal key, tennis ball, book, pen, coin, glass marble, leaf, feather, hammer, egg, orange, plastic cup, wooden block, bolt, rubber eraser, walnut, rubber ball, ceramic mug}
{table edge, held by a hand, tree branch, shelf, window ledge, rooftop, ladder rung, cliff edge}
{motion, direction, spatial, causal}{25}

\categorycard{Gravity\_Moon}{Mechanics}{mechanics}
{Object falls under the Moon's weak gravity ($\sim\!1.6\,\text{m}/\text{s}^2$), noticeably slower than on Earth.}
{Lunar gravity is $\sim\!1/6$ of Earth's; trajectory is parabolic with stretched timescale.}
{astronaut helmet, wrench, feather, stone, ping-pong ball, apple, golf ball, metal bolt, sand bag, rubber ball, glass marble, coin, pen, wooden cube, hammer}
{lunar surface, moon crater, low-gravity habitat, astronaut's hand, space station floor}
{motion, direction, spatial, causal}{20}

\categorycard{Collision\_Elastic}{Mechanics}{mechanics}
{Two objects collide and rebound with kinetic energy conserved; objects bounce apart.}
{Momentum and kinetic energy are both conserved; objects deform momentarily and rebound at predictable velocities.}
{rubber ball, tennis ball, basketball, billiard ball, ping-pong ball, steel ball bearing, elastic ring, football, superball, golf ball, handball, inflated balloon, hollow plastic ball, hard rubber cube, bowling ball}
{hard wall, concrete floor, wooden floor, racket surface, glass surface, another ball, metal plate}
{motion, direction, interaction, causal}{25}

\categorycard{Collision\_Inelastic}{Mechanics}{mechanics}
{Objects collide and stick together or deform permanently; kinetic energy is lost.}
{Momentum is conserved but kinetic energy is not; objects deform, stick, or shatter, and relative velocity decreases.}
{wet clay, soft dough, mud ball, ripe tomato, rice cake, putty, overripe banana, raw egg, soft cheese, cookie dough, wet sand ball, mashed potato ball, playdough cube, very ripe peach, soft butter block}
{wall, hard floor, wooden board, glass window, shirt fabric, another clay lump}
{motion, shape, interaction, causal}{25}

\categorycard{Elastic\_Deformation}{Mechanics}{mechanics}
{Material is compressed or stretched and returns to its original shape when force is removed.}
{Hooke's law: restoring force proportional to displacement; elastic strain energy is fully recovered.}
{stress ball, rubber band, sponge, spring, trampoline, foam pillow, rubber eraser, silicone mold, elastic hair tie, bungee cord, rubber glove, inflated balloon, memory foam, soft rubber cube, latex tube}
{hand grip, between two fingers, on a flat surface, hanging from a hook, pressed against a wall}
{shape, motion, material, causal}{25}

\categorycard{Inelastic\_Deformation}{Mechanics}{mechanics}
{Material is deformed beyond its elastic limit and retains the new shape permanently.}
{Plastic deformation: stress exceeds yield strength, causing permanent rearrangement of internal structure without recovery.}
{clay lump, playdough, soft lead cylinder, wax block, putty, warm butter stick, wet sand mound, foil ball, soft cheese block, raw dough ball, modeling clay ring, beeswax slab, overripe avocado, mud cake, soft chocolate block}
{hand, flat surface, mold, rolling pin, finger press, hammer strike, wooden stamp}
{shape, material, motion, causal}{20}

\categorycard{Friction\_High}{Mechanics}{mechanics}
{Object slides over a high-friction surface and decelerates rapidly, stopping in a short distance.}
{Kinetic friction $F=\mu_k N$; high $\mu$ yields large deceleration, with possible heat or wear.}
{wooden block, book, cardboard box, eraser, brick, sandpaper block, rubber-soled shoe, rough ceramic tile, hockey puck, foam block, metal plate with rough coating, coarse stone, fabric bag, rubber mat, wicker basket}
{rough concrete, thick grass, sandpaper surface, wooden table, carpet, gravel path, rubber mat, coarse stone floor}
{motion, direction, spatial, causal}{25}

\categorycard{Friction\_Low}{Mechanics}{mechanics}
{Object glides over a low-friction surface and travels a long distance before stopping.}
{Low coefficient of friction allows kinetic energy to persist; sliding distance much greater than on rough surfaces.}
{curling stone, Lego block, flat stone, ice cube, hockey puck, polished metal disc, glass marble, air hockey puck, Teflon-coated block, smooth plastic toy, waxed wooden block, butter stick, soap bar, frozen fish fillet, steel washer}
{ice surface, glass table, polished marble floor, waxed wooden floor, wet tile, oiled metal plate, playground slide, Teflon sheet}
{motion, direction, spatial, causal}{25}

\categorycard{Pendulum}{Mechanics}{mechanics}
{A mass suspended from a fixed pivot swings back and forth in periodic motion.}
{Simple pendulum: $T=2\pi\sqrt{L/g}$, mass-independent; potential and kinetic energy alternate; amplitude decays slowly.}
{metal ball, wooden sphere, stone weight, clock pendulum bob, steel nut on string, glass ball, ceramic bead, rubber ball, lead weight, brass sphere, acorn, walnut on string, small lantern, crystal pendant, metal ring}
{string attached to ceiling, rod pivot, grandfather clock, playground swing, hanging from door frame, outdoor tree branch}
{motion, direction, spatial, causal}{20}

\categorycard{Rolling\_Motion}{Mechanics}{mechanics}
{A round object rolls along a surface, exchanging translational and rotational kinetic energy.}
{Rolling without slipping: $v=\omega r$; total $\text{KE}=\tfrac{1}{2}mv^2+\tfrac{1}{2}I\omega^2$; mass distribution affects roll rate.}
{bowling ball, basketball, marble, coin, cylinder can, tennis ball, wooden barrel, metal pipe, wheel, soup can, empty paint roller, tire, log, wooden spool, hollow ball}
{inclined ramp, flat floor, grass, carpet, kitchen counter, bowling lane, concrete slope}
{motion, direction, spatial, causal}{25}

\categorycard{Projectile\_Motion}{Mechanics}{mechanics}
{An object is launched and follows a parabolic arc under gravity with no propulsion mid-flight.}
{Horizontal velocity constant, vertical velocity grows at $g$; range depends on launch angle and speed.}
{baseball, basketball, football, tennis ball, stone, frisbee, javelin, shuttlecock, golf ball, paper airplane, rubber ball, coin, water balloon, beanbag, snowball}
{open field, gymnasium, basketball court, garden, rooftop, stadium, beach, playground}
{motion, direction, spatial, causal}{25}

\categorycard{Structure\_Failure}{Mechanics}{mechanics}
{A structure collapses under excessive load, impact, or material fatigue.}
{Stress exceeds ultimate strength; failure propagates from the weakest point and pieces fall under gravity.}
{wooden bridge, sandcastle, card tower, Jenga tower, snow fort, styrofoam arch, clay pillar, dominoes structure, matchstick bridge, balsa wood frame, plaster column, foam house, sugar-cube tower, cracker wall}
{table, sand, floor, ground, platform, display surface}
{shape, motion, spatial, causal}{20}

\paragraph{Fluids.}

\categorycard{Buoyancy\_Float}{Fluid}{fluid}
{Object placed in a fluid displaces its weight in fluid and floats at the surface.}
{Archimedes' principle: $F_b=\rho_{\text{fluid}}V_{\text{disp}}g$; object floats when its average density is below fluid density.}
{rubber duck, sponge, ice cube, wooden toy, plastic bottle, cork, foam block, ping-pong ball, toy boat, wax candle, apple, coconut, empty plastic cup, styrofoam sphere, balsa wood plank, hollow metal ball, pumice stone, sealed empty bottle, wooden log, inflated balloon}
{bathtub, glass tank, swimming pool, ocean, lake, bucket, fish tank, river, puddle}
{motion, spatial, material, causal}{25}

\categorycard{Buoyancy\_Sink}{Fluid}{fluid}
{Object placed in a fluid is denser than the fluid and sinks to the bottom.}
{When average density exceeds fluid density, net force is downward and the object accelerates toward the bottom.}
{iron block, heavy rock, steel spoon, anchor, ceramic cup, metal coin, glass marble, lead weight, bronze statue, copper pipe, stone pebble, iron nail, metal bolt, ceramic tile, dense rubber ball, concrete cube, sand-filled bag, zinc washer, cast-iron pan, brass weight}
{glass tank, bathtub, swimming pool, bucket, fish tank, ocean floor, clear container, aquarium}
{motion, spatial, material, causal}{25}

\categorycard{Viscous\_Flow}{Fluid}{fluid}
{A high-viscosity fluid flows slowly and resists deformation, spreading sluggishly.}
{Navier--Stokes for high-viscosity Newtonian fluids; flow rate $\propto 1/\mu$; thick fluids hold shape and resist spreading.}
{honey, condensed milk, thick oil, molasses, blueberry jam, peanut butter, caramel sauce, corn syrup, tar, maple syrup, thick batter, ketchup, chocolate sauce, lava, tree sap, gel, whipped cream, thick glue, paint, yogurt}
{toast, plate, pancakes, spoon, inclined surface, funnel, glass jar, bowl, bread slice, waffle}
{motion, shape, spatial, causal}{25}

\categorycard{Splash}{Fluid}{fluid}
{An object impacts a liquid surface, displacing fluid upward into a crown or column.}
{Impact converts KE into surface waves and ejected droplets; crown height $\propto$ impact speed; droplet size depends on surface tension.}
{strawberry, grape, cherry, marble, small stone, lemon slice, ice cube, toy ball, metal coin, sugar cube, rubber duck, tomato, blueberry, walnut, apple slice, strawberry cream, small rubber ball, ping-pong ball, glass bead, small steel ball}
{glass of water, glass of milk, glass of juice, shallow pool, puddle, aquarium, swimming pool}
{motion, spatial, shape, causal}{25}

\categorycard{Surface\_Tension}{Fluid}{fluid}
{Liquid surface acts like a stretched membrane; small objects rest on it and droplets form spherical shapes.}
{Surface tension $\gamma$ from cohesive forces at the liquid--air interface; droplet pressure $\Delta P=2\gamma/r$.}
{water droplet, soap bubble, water strider insect, needle, paper clip, small coin, thin metal wire, razor blade, drop of oil, mercury droplet, dew droplet, raindrop, drop of food coloring, small leaf, thin plastic disc}
{water surface, glass of water, soap solution, spider web, window glass, petri dish, calm pond}
{shape, spatial, interaction, causal}{20}

\categorycard{Capillary\_Action}{Fluid}{fluid}
{Liquid rises or spreads through narrow spaces against gravity due to adhesion and surface tension.}
{Capillary rise $h=2\gamma\cos\theta/(\rho g r)$; narrower tubes produce higher rise as adhesive forces overcome gravity.}
{paper towel, thin glass tube, celery stalk, white flower stem, cotton wick, sponge, chalk stick, sugar cube, thin straw, capillary tube, brick, cloth strip, dried pasta, thin wooden stick, filter paper strip}
{colored water, ink solution, water with food dye, petri dish, beaker, shallow tray, glass of water}
{motion, direction, spatial, causal}{20}

\categorycard{Liquid\_Layer}{Fluid}{fluid}
{Two immiscible liquids of different densities separate into distinct horizontal layers.}
{Immiscible fluids stratify by density; mixing energy dissipates and layers re-separate if undisturbed.}
{cooking oil, water, honey, corn syrup, rubbing alcohol, dish soap, maple syrup, milk, vinegar, food-colored water, baby oil, glycerine, lamp oil, light mineral oil}
{tall glass cylinder, mason jar, test tube, clear plastic bottle, fish tank, transparent jug}
{spatial, appearance/texture, material, causal}{20}

\categorycard{Liquid\_Mixing}{Fluid}{fluid}
{Two miscible liquids are combined, diffusing and blending into a uniform mixture.}
{Miscible liquids mix by diffusion and convection; without stirring mixing is slow, with agitation it is rapid.}
{red food dye, blue food dye, yellow food dye, milk, coffee, juice, vinegar, salt water, warm water, cold water, soap solution, sugar syrup, green tea, black ink drop, lemon juice}
{glass of water, bowl, clear container, mixing cup, aquarium, transparent pitcher}
{appearance/texture, spatial, motion, causal}{25}

\categorycard{Granular\_Flow}{Fluid}{fluid}
{A collection of discrete solid particles flows and piles up like a fluid under gravity.}
{Granular media exhibit solid-like rest and fluid-like flow; angle of repose sets pile shape; flow rate depends on particle size and geometry.}
{sand, rice, sugar, salt, coffee beans, dry soil, gravel, small pebbles, wheat flour, cornmeal, bird seed, lentils, sesame seeds, pasta pieces, glass beads, mustard seeds, crushed dry leaves, aquarium gravel, fine sawdust, coarse salt}
{hand, funnel, hourglass, glass container, flat surface, pile on ground, inclined ramp, jar, bowl, transparent tube}
{motion, shape, spatial, causal}{25}

\categorycard{Vortex}{Fluid}{fluid}
{A rotating fluid forms a funnel-shaped vortex around a low-pressure core.}
{Angular-momentum conservation makes fluid spin faster near the center; centripetal acceleration balances pressure gradient.}
{water in a drain, stirred cup of tea, tornado, dust devil, whirlpool in a bathtub, stirred paint pot, milk swirling in coffee, whisk in batter, stirred aquarium, draining sink}
{bathtub, sink drain, large glass, bucket, mixing bowl, fish tank, clear container}
{motion, shape, spatial, causal}{20}

\paragraph{Materials.}

\categorycard{Fracture}{Material}{materials}
{A brittle object breaks into fragments when stress exceeds its fracture toughness.}
{Brittle fracture: crack propagates rapidly once $K_{Ic}$ is reached; no plastic flow, shards radiate from impact.}
{wine glass, ceramic plate, glass bottle, mirror, ceramic vase, porcelain cup, glass window, clay pot, glass fish bowl, tile, glass sphere, china bowl, glass panel, brittle plastic ruler, crystal ornament}
{hard floor, concrete surface, stone counter, wooden floor, brick wall, metal anvil, another hard object}
{shape, motion, spatial, causal}{25}

\categorycard{Slicing}{Material}{materials}
{A sharp blade moves through an object, cutting it cleanly into two or more pieces.}
{Stress concentration at the blade edge exceeds shear strength; cut quality depends on sharpness, force, and hardness.}
{bread loaf, carrot, apple, cucumber, tomato, rope, cheese block, banana, clay block, soft wood, marshmallow, butter block, watermelon, mushroom, raw chicken breast, celery stalk, onion half, sponge cake, eggplant, zucchini}
{cutting board, kitchen counter, flat surface, butcher block, marble slab}
{shape, interaction, motion, causal}{25}

\categorycard{Tearing}{Material}{materials}
{A soft or fibrous material is pulled apart, propagating a crack or tear along the material.}
{Crack-tip stress concentration drives propagation in ductile/fibrous materials; fibers stretch then break sequentially.}
{newspaper, paper sheet, paper towel, plastic bag, fabric cloth, rubber sheet, thin leather, foil sheet, tissue paper, wet paper, duct tape, thin plastic film, rice paper, nori sheet, cotton fabric}
{two hands, hands and fixed edge, tearing machine, sharp corner, nail, flat surface}
{shape, motion, interaction, causal}{20}

\categorycard{Cloth\_Simulation}{Material}{materials}
{Flexible fabric moves realistically under wind, gravity, or contact forces.}
{Cloth modeled as a mass--spring system; gravity pulls down, tension resists stretch, aerodynamic forces cause flutter.}
{silk flag, curtain, tablecloth, scarf, flag, bed sheet, cape, paper flag, light dress fabric, handkerchief, gossamer curtain, towel, canvas sail, ribbon, thin plastic sheet, parachute}
{flagpole, window frame, chair back, clothesline, wind outdoors, fan indoors, falling through air}
{motion, shape, spatial, causal}{25}

\categorycard{Explosion}{Material}{materials}
{Rapid release of energy causes a pressure wave, fragmentation, and outward dispersion of debris.}
{Exothermic reaction or pressure overload releases energy faster than dissipation, creating a shock and ejecting fragments radially.}
{inflated balloon, gasoline drum, pressurized container, firecracker, water balloon, plastic soda bottle, popcorn kernel, spray can, overinflated tire, bubble underwater, sealed jar heated, compressed-gas cylinder, party popper, exploding cake, pressure cooker failure}
{open space, indoor room, underwater, concrete surface, metal container, lab bench}
{motion, spatial, shape, causal}{20}

\categorycard{Smoke}{Material}{materials}
{Small particles or vapor rise and disperse in air, forming visible plumes.}
{Buoyancy drives hot smoke upward ($\rho_{\text{smoke}}<\rho_{\text{air}}$); diffusion and turbulent mixing spread the plume.}
{lit candle, incense stick, blown-out candle, cup of hot tea, cigarette, campfire embers, hot food, dry ice, smoke machine, chimney, burning match, hot iron, fogger device, extinguished fire, steam pipe}
{indoor room, outdoor environment, dim room, kitchen, fire pit, stage, open air}
{motion, direction, shape, causal}{25}

\categorycard{Combustion}{Material}{materials}
{A material burns, producing visible flame, smoke, light, and heat.}
{Rapid oxidation: $\text{fuel}+\text{O}_2\rightarrow\text{CO}_2+\text{H}_2\text{O}+\text{heat}$; flame color reflects temperature and fuel.}
{candle wick, wood log, paper sheet, match stick, dried leaf, twig, birthday candle, torch, alcohol lamp, oil lamp, campfire log, sparkler, newspaper, cardboard, wax tablet}
{fireplace, campfire pit, lab bench, candle holder, outdoor setting, metal bowl, fire pit}
{appearance/texture, motion, shape, causal}{20}

\categorycard{Chain\_Reaction}{Material}{materials}
{An initial perturbation triggers a cascade of sequential events, each causing the next.}
{Each element stores potential energy that is transferred upon impact; total energy propagates down the chain.}
{dominoes, mousetraps, billiard balls in line, falling books, chain of coins, Newton's cradle, lined-up toy blocks, stacked dominoes, bowling pins, linked magnets, falling card stack, toy train collision, cascading water buckets, toppling columns, matchstick chain}
{flat table, wooden floor, ramp, gymnasium floor, outdoor surface, display table}
{motion, number, direction, causal}{20}

\paragraph{Thermal.}

\categorycard{Melting}{Thermal}{thermal}
{A solid gradually transforms into a liquid as temperature rises above its melting point.}
{Latent heat absorbed at $T_m$ breaks crystalline bonds; volume may shift slightly and shape collapses.}
{ice cube, candle wax, butter, snow patch, ice cream, chocolate bar, wax crayon, lead block, paraffin candle, ice pop, frozen yogurt, wax seal, lard block, ice sculpture, gelatin block}
{warm plate, heated pan, direct sunlight, microwave plate, warm water bath, room temperature counter, hot surface}
{shape, appearance/texture, motion, causal}{25}

\categorycard{Freezing}{Thermal}{thermal}
{A liquid cools below its freezing point and solidifies, changing shape and texture.}
{Latent heat is released as the liquid loses thermal energy; crystal structure forms (water expands $\sim$9\%).}
{water in tray, juice in mold, wet cloth, puddle, lake surface, dripping water, water in cup, mercury thermometer liquid, fruit juice, coconut water, lemon juice, herbal tea, smoothie, sports drink, broth}
{freezer interior, cold winter outdoor, ice bath, refrigerator shelf, liquid nitrogen container, cold metal surface, outdoor frost}
{shape, appearance/texture, material, causal}{20}

\categorycard{Evaporation}{Thermal}{thermal}
{Surface molecules escape to vapor, gradually reducing liquid volume.}
{Surface evaporation rate scales with temperature, area, and vapor-pressure deficit; latent heat is absorbed.}
{water puddle, wet cloth, water droplets on leaf, glass of water, wet pavement, ink droplet, rubbing alcohol drop, acetone pool, wet sand, water on hot pan, morning dew, wet shoe sole, brine droplet, perfume drop, water on glass surface}
{hot surface, outdoor sunlight, warm room, heated plate, dry air environment, desert setting}
{appearance/texture, size, motion, causal}{20}

\categorycard{Boiling}{Thermal}{thermal}
{Liquid heated to its boiling point forms vapor bubbles that rise and burst at the surface.}
{Vapor pressure equals atmospheric pressure at $T_b$; nucleate bubbles form on the heated surface.}
{water in pot, milk in saucepan, oil in pan, soup, broth, tea kettle water, sugar syrup, salt water, egg in boiling water, pasta in pot, jam in pan, hot chocolate, caramel mixture, stew liquid, gravy}
{stove burner, hot plate, campfire, induction cooker, lab hotplate, fondue pot}
{appearance/texture, motion, spatial, causal}{20}

\categorycard{Thermal\_Expansion}{Thermal}{thermal}
{A material expands visibly as its temperature increases due to increased molecular vibration.}
{Linear expansion $\Delta L=\alpha L_0\Delta T$; gases expand most, metals less.}
{metal rod, balloon over bottle neck, bimetallic strip, thermometer liquid column, rubber ball, air in sealed syringe, plastic bottle with air, metal lid on jar, metal ring, coiled spring, wire on posts, sealed air balloon}
{hot water bath, heat gun, boiling water, open flame, oven, sunlight, heating pad}
{size, shape, motion, causal}{20}

\categorycard{Thermal\_Convection}{Thermal}{thermal}
{Fluid circulates as heated portions rise and cooled portions sink, transferring heat.}
{Density decreases with temperature; Rayleigh--B\'enard cells form in heated fluid layers.}
{colored hot water, lava lamp fluid, hot cocoa, ink in warm water, heated oil, glycerine with dye, water with food coloring, heated aquarium water, boiling soup, warm sugar solution}
{transparent container, lava lamp, glass bowl, aquarium, heated beaker, lab flask}
{motion, direction, spatial, causal}{20}

\categorycard{Crystallization}{Thermal}{thermal}
{Dissolved or molten material forms ordered crystal structures as it cools or concentrates.}
{Crystal nucleation and growth: ions/molecules arrange into a minimum-energy lattice; rate depends on supersaturation.}
{salt solution, sugar solution, borax solution, copper sulfate solution, alum solution, cooling chocolate, honey crystallizing, cooling wax, cooling metal cast, rock candy solution, frost on window}
{petri dish, jar, glass container, cooling surface, cold window, refrigerator shelf}
{shape, appearance/texture, size, causal}{20}

\categorycard{Condensation}{Thermal}{thermal}
{Water vapor cools below dew point and condenses into liquid droplets on a cold surface.}
{Surface temperature below dew point releases latent heat; droplets grow, coalesce, and run.}
{cold glass, cold metal can, bathroom mirror, cold window glass, cold pipe, ice-cold bottle, refrigerator exterior, cold stone surface, cold car windshield, cold metal plate}
{warm humid room, bathroom, outdoor morning, refrigerator exterior, kitchen, greenhouse}
{appearance/texture, size, spatial, causal}{20}

\paragraph{Electromagnetism.}

\categorycard{Reflection}{Electromagnetism}{electromagnetism}
{Light or image bounces off a reflective surface at equal angles of incidence and reflection.}
{Law of reflection: $\theta_i=\theta_r$; specular surfaces produce mirror images, diffuse surfaces scatter light.}
{dancer in mirror, person waving, car on wet road, sunlight on lake, man brushing teeth, ball rolling to mirror, candle flame in mirror, hand gesturing, text on mirror, laser beam on mirror, lamp reflected in polished floor, face in spoon reflection, sky in puddle, building in glass facade, fire reflected in water}
{flat mirror, glass window, still lake, wet road, polished metal, calm water pool, bathroom mirror, shop window, chrome surface}
{direction, spatial, motion, appearance/texture}{25}

\categorycard{Refraction}{Electromagnetism}{electromagnetism}
{Light bends at the boundary between media of different refractive indices, displacing apparent positions.}
{Snell's law: $n_1\sin\theta_1=n_2\sin\theta_2$; objects in water appear shifted; lenses and prisms exploit refraction.}
{spoon in water, straw in glass, pencil in water, banana behind glass, coin at bottom of bowl, fish in aquarium, fork in water, pen in glass of water, laser beam entering water, prism splitting white light, magnifying glass focusing light, straw appearing at angle, ruler in water, chopstick in glass}
{glass of water, aquarium, glass prism, lens setup, fish tank, clear water bowl, glass cylinder}
{shape, appearance/texture, spatial, direction}{25}

\categorycard{Shadow}{Electromagnetism}{electromagnetism}
{An opaque object blocks a light source, casting a shadow whose shape and length depend on geometry.}
{Shadow boundary traced by tangent rays; shadow length $=h\cdot D/d$ for source--object distance $D$ and object--screen distance $d$.}
{person walking, rotating basketball, Lego blocks, tree branch, moving hand puppet, swinging pendulum, passing car, rotating fan blade, moving cloud, candle with obstacle, hand blocking flashlight, flying bird silhouette, plant on windowsill, book under lamp, spinning top}
{sunlit wall, ground under sunlight, dim room with lamp, screen behind object, painted wall, sandy ground, white sheet, whiteboard}
{shape, direction, spatial, motion}{25}

\categorycard{Magnetism}{Electromagnetism}{electromagnetism}
{Magnetic objects attract or repel each other, and ferromagnetic materials are drawn to magnets.}
{Magnetic force $F\propto$ field gradient on dipoles; like poles repel and unlike poles attract; iron and nickel are strongly attracted.}
{bar magnet, horseshoe magnet, neodymium magnet, iron nail, steel paperclip, iron filing, compass needle, steel key, iron washer, steel ball bearing, iron powder, refrigerator magnet, magnetic toy, metal coin, iron chain link}
{table surface, air, paper substrate, glass table, sand surface, hanging from string, resting on wood}
{motion, direction, interaction, causal}{20}

\categorycard{Electrostatics}{Electromagnetism}{electromagnetism}
{Charged objects attract neutral or oppositely charged objects via the electrostatic force.}
{Coulomb's law $F=kq_1q_2/r^2$; triboelectricity charge generated by contact and friction (e.g., balloon attracts paper).}
{inflated balloon, plastic comb, PVC rod, wool fabric, bits of paper, hair strands, small styrofoam beads, thin aluminum foil strips, charged amber rod, puffed rice grains, shredded paper, thin plastic strips, pepper grains, thin string, dried flower petals}
{table top, air, hair surface, paper surface, near balloon, near charged rod, enclosed jar}
{motion, direction, interaction, causal}{20}

\paragraph{Chemistry and Material Interaction.}

\categorycard{Diffusion}{Chemistry}{chemistry}
{Particles or molecules spread from high to low concentration until uniformly distributed.}
{Fick's law: flux $\propto -D\nabla C$; Brownian motion drives equilibrium; faster in gases, slower in liquids.}
{drop of ink in water, food dye drop, perfume spray, drop of milk in tea, potassium permanganate crystal, drop of colored dye, smoke in still air, drop of cream in coffee, gas in sealed chamber, colored sugar drop in water, red wine drop in water, turmeric in water, drop of oil in alcohol}
{glass of still water, petri dish, transparent container, still air, closed jar, aquarium}
{motion, appearance/texture, spatial, causal}{20}

\categorycard{Dissolution}{Chemistry}{chemistry}
{A solute disperses into a solvent at the molecular level, forming a homogeneous solution.}
{Solvation: solvent molecules surround solute species; rate increases with temperature, stirring, and surface area.}
{salt crystal, sugar cube, effervescent tablet, bath bomb, instant coffee granules, hard candy, Alka-Seltzer tablet, vitamin C tablet, chalk piece in acid, baking soda, copper sulfate crystal, food dye tablet, bouillon cube, aspirin tablet, dry gelatin powder}
{glass of water, warm water bowl, cold water glass, vinegar solution, juice, clear container}
{appearance/texture, size, motion, causal}{20}

\categorycard{Rust\_Oxidation}{Chemistry}{chemistry}
{Iron or steel reacts with oxygen and moisture, forming reddish-brown iron oxide over time (time-lapse).}
{Electrochemical corrosion $\text{Fe}+\text{O}_2+\text{H}_2\text{O}\rightarrow\text{Fe}_2\text{O}_3\!\cdot\! n\text{H}_2\text{O}$; porous rust permits further oxidation.}
{iron nail, steel nail, iron plate, steel wool, iron chain, steel bolt, iron wire mesh, cast iron pan, steel washer, iron railing, steel screw, iron beam section, steel fence post, iron pipe section, steel file}
{salt water, moist air, outdoor rain exposure, humid room, salt spray, wet soil}
{appearance/texture, size, material, causal}{20}

\categorycard{Color\_Mixing}{Chemistry}{chemistry}
{Two or more colored substances combine, producing a new mixed color.}
{Subtractive (CMY) mixing absorbs complementary wavelengths; additive (RGB) mixing increases brightness.}
{red paint, blue paint, yellow paint, red food dye, blue food dye, yellow dye solution, watercolor paints, red ink, blue ink, green dye, white paint, black paint, cyan dye, magenta dye, colored powder}
{palette, mixing bowl, glass of water, paper surface, paint tray, clear container, canvas}
{appearance/texture, motion, interaction, causal}{25}

\categorycard{Adsorption}{Chemistry}{chemistry}
{Molecules adhere to a material's surface, removing them from solution or gas phase.}
{Surface adsorption via van der Waals or chemical bonds; capacity depends on surface area and affinity.}
{activated charcoal, sponge absorbing liquid, paper towel, silica gel, zeolite beads, clay powder, iron filings attracting oil, dry cloth in colored water, coconut shell charcoal, porous ceramic bead, coffee filter paper}
{colored water, ink solution, dye bath, oil-contaminated water, clear container, transparent beaker}
{appearance/texture, motion, interaction, causal}{20}

\categorycard{Precipitation}{Chemistry}{chemistry}
{Two solutions react to form an insoluble solid that falls out of solution.}
{Solubility product exceeded ($[\text{A}^+][\text{B}^-]>K_{sp}$); precipitate nucleates, grows, and settles.}
{silver nitrate and sodium chloride, barium chloride and sodium sulfate, lead nitrate and sodium iodide, calcium chloride and sodium carbonate, iron(III) chloride and sodium hydroxide, copper sulfate and sodium hydroxide, lime water and CO$_2$, milk of magnesia forming, calcium carbonate forming, zinc hydroxide forming}
{test tube, beaker, clear container, glass mixing vessel, transparent flask}
{appearance/texture, motion, spatial, causal}{20}

\categorycard{Foam\_Bubble}{Chemistry}{chemistry}
{Gas trapped in liquid forms bubbles or foam; bubbles rise, coalesce, and burst.}
{Bubble pressure $\Delta P=4\gamma/r$; foam drains over time; surfactants stabilize the structure.}
{soap foam in sink, beer foam, latte milk foam, whipped cream, shaving cream, bubble bath foam, carbonated water bubbles, lemon juice with baking soda, hand soap foam, fizzy drink bubbles, sea foam, foam from waterfall, bubbles in boiling milk, hydrogen peroxide foam, foam from soap squeeze}
{sink basin, glass, bowl, counter surface, cup, bucket, transparent container}
{shape, motion, size, causal}{20}

\subsection{Structured Physical Priors Construction}
\label{appen:priors_construction}
For each scene in \phymetric{}, we attach a \emph{structured physical prior} that anchors the VLM judges to the governing law and its expected visual signatures before any question is asked.
To avoid the situation where VLM judges obtains homologous LLMs priors, we utilize four different LLMs to provide physical priors to conduct our ablation experiment in Section~\ref{subsec:ablation}.

\subsubsection{LLMs backbones}

The structured physical priors in \phymetric{} are generated using a frozen text-based LLM. 
To validate that our evaluation framework is model-agnostic and free from specific LLM-based prior shortcuts, we employ four representative LLMs to generate these structured physical priors. 
This selection includes models with different architectural lineages (Qwen, Mistral, and DeepSeek) and reasoning capabilities.
As detailed in Table~\ref{tab:kb_models}, \textbf{Qwen3.5-9B} is used as the default generator. 
For our ablation studies, we incorporate \textbf{Qwen2.5-VL-7B-Instruct} (text-mode), \textbf{DeepSeek-R1-Distill-Llama-8B}, and \textbf{Mistral-7B-Instruct-v0.3}. 
This diverse ensemble allows us to investigate the sensitivity of VLM judges to varying qualities of physical reasoning and ensures the generalizability of our proposed physical prior anchoring.

\begin{table}[ht]
\centering
\small
\setlength{\tabcolsep}{6pt}
\caption{LLM backbones used for structured physical prior generation. All models are run text-only and the default backbone for the released benchmark is \textbf{Qwen3.5-9B} in text mode.}
\label{tab:kb_models}
\begin{tabular}{lll}
\toprule
\textbf{Model} & \textbf{Role} & \textbf{Notes} \\
\midrule
Qwen3.5-9B (thinking)     & Default backbone         & Thinking suppressed via chat-template flag \\
Qwen2.5-VL-7B-Instruct  & Ablation & Text-only mode; \texttt{enable\_thinking=False} \\
DeepSeek-R1-Distill-Llama-8B     & Ablation         & Strips \texttt{<think>...</think>}; budget 16{,}384 tokens \\
Mistral-7B-Instruct-v0.3 & Ablation         & Drop-in via \texttt{TextLLM} wrapper \\
\bottomrule
\end{tabular}
\end{table}

\subsubsection{System prompt}

The prompt is constructed from five scene-level fields: the physical phenomenon (namely physical category), scene name, scene caption, the union of question dimensions across the eight binary questions in that scene, and the benchmark video metadata (including video duration, fps, and resolution).
To facilitate question-aware frame sampling, we define three deterministic reference time points at 25\%, 50\%, and 83\% of the total duration. In our 120-frame clips, these points correspond to frames 30, 60, and 100, ensuring that the early, middle, and late physical states described in the prior align precisely with the frames sampled by the VLM judge during evaluation.

\begin{tcolorbox}[
  enhanced, breakable, colback=gray!5, colframe=black!55,
  title=\textbf{System Prompt} (\texttt{PHYSICS\_PRIORS}),
  fonttitle=\bfseries, left=4pt, right=4pt, top=2pt, bottom=2pt,
  boxrule=0.6pt, arc=2pt
]
\small\ttfamily
You are an expert physicist and video generation evaluator. \\
Your task is to produce a concise but comprehensive knowledge base that will later help a vision-language model judge whether a short generated video correctly depicts a physical scene. Keep any internal reasoning brief. \\
Respond ONLY with a valid JSON object, no markdown, and no extra text.
\end{tcolorbox}

\begin{tcolorbox}[
  enhanced, breakable, colback=blue!3, colframe=blue!88!black!55,
  title=\textbf{User Prompt Template} (\texttt{PHYSICS\_PRIORS}),
  fonttitle=\bfseries, left=4pt, right=4pt, top=2pt, bottom=2pt,
  boxrule=0.6pt, arc=2pt
]
\scriptsize\ttfamily
Physical phenomenon category: \{phenomenon\} \\
Scene name: \{scene\_name\} \\
Scene caption (the prompt given to the video model): \\
\hspace*{1em}"\{scene\_caption\}" \\[2pt]
Benchmark video spec: \{duration\_s\} seconds, \{fps\} fps, \{total\_frames\} frames total. \\
Three reference time-points (benchmark frame indices): \\
\hspace*{1em}- Early : frame \{ref\_early\} (\textasciitilde\{t\_early\} s) \\
\hspace*{1em}- Mid   : frame \{ref\_mid\}   (\textasciitilde\{t\_mid\} s) \\
\hspace*{1em}- Late  : frame \{ref\_late\}  (\textasciitilde\{t\_late\} s) \\[2pt]
Question dimensions present in this scene: \{dimensions\} \\[4pt]
Generate a structured physical prior JSON with exactly these keys: \\[2pt]
\{ \\
\hspace*{1em}"physical\_law": "<1-3 sentences: the governing physical law, key conditions, \\
\hspace*{4em}and what drives the event>", \\
\hspace*{1em}"expected\_sequence": \{ \\
\hspace*{2em}"early": "<physical state at \textasciitilde\{t\_early\} s: what has started, what force/process is acting>", \\
\hspace*{2em}"mid":   "<physical state at \textasciitilde\{t\_mid\} s: how the event has progressed, what is changing>", \\
\hspace*{2em}"late":  "<physical state at \textasciitilde\{t\_late\} s: approaching final state, what outcome is visible>" \\
\hspace*{1em}\}, \\
\hspace*{1em}"physics\_signatures": [ \\
\hspace*{2em}"<MUST-observe consequence 1: a physically necessary change/motion/interaction \\
\hspace*{4em}that the law dictates --- describe the phenomenon type, NOT a specific pixel \\
\hspace*{4em}position or frame number>", \\
\hspace*{2em}"<MUST-observe consequence 2: what would be impossible or implausible if this \\
\hspace*{4em}law is violated>", \\
\hspace*{2em}"<MUST-observe consequence 3: the key state transition that distinguishes \\
\hspace*{4em}physically correct from incorrect behavior>" \\
\hspace*{1em}], \\
\hspace*{1em}"common\_failures": [ \\
\hspace*{2em}"<typical video-generation failure 1: how the model usually gets this wrong>", \\
\hspace*{2em}"<failure 2>", \\
\hspace*{2em}"<failure 3>" \\
\hspace*{1em}], \\
\hspace*{1em}"evaluation\_hints": \{ \\
\hspace*{2em}\{dimension\_hints\_template\} \\
\hspace*{1em}\} \\
\} \\[4pt]
Rules for physics\_signatures: \\
\hspace*{1em}- Describe observable *physical phenomena* (acceleration, deformation, state change, \\
\hspace*{4em}trajectory type). \\
\hspace*{1em}- Do NOT prescribe exact pixel positions, object locations, or specific frame timestamps. \\
\hspace*{1em}- Each item should be something a viewer can confirm from any reasonable depiction \\
\hspace*{4em}of the correct physics. \\[2pt]
Be concrete and reference the actual objects in the scene caption.
\end{tcolorbox}

\subsection{VLMs Evaluation Details}
\label{appen:vlm_eval}
\subsubsection{System and user prompt}

For every question, the VLM judge receives (i) a fixed system message specifying the judging contract, (ii) a small set of frames sampled at the question's \texttt{frame\_indices}, and (iii) a user message that splices together the scene caption, the video specification, the structured physical prior, and the binary question.
The judge is required to generate \texttt{ANSWER}, \texttt{CONFIDENCE}, and \texttt{REASON} on three lines so that downstream evaluation can extract a $\{$yes,~no$\}$ answer with a calibrated confidence.

\begin{tcolorbox}[
  enhanced, breakable, colback=gray!5, colframe=black!55,
  title=\textbf{System Prompt} (\texttt{VLM\_REASONING}),
  fonttitle=\bfseries, left=4pt, right=4pt, top=2pt, bottom=2pt,
  boxrule=0.6pt, arc=2pt
]
\small\ttfamily
You are a physics-aware video evaluation assistant. You will be shown a sequence of frames from a short AI-generated video, a structured physical prior describing what the video SHOULD depict physically, and a yes/no question about the video content. Reason carefully using the physical knowledge, then give your answer. \\[2pt]
Your response MUST follow this exact format (one line each): \\
ANSWER: YES \\
CONFIDENCE: 0.92 \\
REASON: <one sentence explaining your judgment>
\end{tcolorbox}

\begin{tcolorbox}[
  enhanced, breakable, colback=blue!3, colframe=blue!88!black!55,
  title=\textbf{User Prompt Template} (\texttt{VLM\_REASONING}),
  fonttitle=\bfseries, left=4pt, right=4pt, top=2pt, bottom=2pt,
  boxrule=0.6pt, arc=2pt
]
\scriptsize\ttfamily
Scene caption: "\{scene\_caption\}" \\
Video spec: \{video\_fps\} fps, \{video\_frames\} frames total, \textasciitilde\{duration\} s. Shown frames are proportionally sampled from this video. \\
Frames shown correspond to benchmark time-points: [\{frame\_indices\}] / 120 \\[4pt]
--- STRUCTURED PHYSICAL PRIOR --- \\
Physical law: \{physical\_law\} \\[2pt]
Expected temporal sequence: \\
\hspace*{1em}\textbullet\ Early: \{expected\_sequence.early\} \\
\hspace*{1em}\textbullet\ Mid:   \{expected\_sequence.mid\} \\
\hspace*{1em}\textbullet\ Late:  \{expected\_sequence.late\} \\[2pt]
Physical law signatures (what MUST be observable in a correct video): \\
\hspace*{1em}\textbullet\ \{physics\_signatures[0]\} \\
\hspace*{1em}\textbullet\ \{physics\_signatures[1]\} \\
\hspace*{1em}\textbullet\ \{physics\_signatures[2]\} \\[2pt]
Common generation failures (what to watch for): \\
\hspace*{1em}\textbullet\ \{common\_failures[0]\} \\
\hspace*{1em}\textbullet\ \{common\_failures[1]\} \\
\hspace*{1em}\textbullet\ \{common\_failures[2]\} \\[2pt]
Evaluation hints for this question's dimensions: \\
\hspace*{1em}\textbullet\ [\{dim\}] \{evaluation\_hints[dim]\} \\
--- END STRUCTURED PHYSICAL PRIOR --- \\[4pt]
Question (answer YES or NO): \{question\_text\}
\end{tcolorbox}


\subsubsection{VLMs judges}

We evaluate three VLM judges that span the open-source and closed-source frontier: two open-source Qwen-family checkpoints at different parameter scales, and a closed-source GPT model.
All judges receive the same prompt and video input described above.

\begin{table}[h]
\centering
\small
\setlength{\tabcolsep}{6pt}
\caption{VLM judges used in \phymetric{}. ``Default'' marks the judge whose scores are reported in~\autoref{tab:results_full} and~\autoref{tab:results_subset}, and the others are used for cross-evaluator robustness checks in~\autoref{subsec:ablation}).}
\label{tab:vlm_judges}
\begin{tabular}{lllcl}
\toprule
\textbf{Judge} & \textbf{Family} & \textbf{Source} & \textbf{Params} & \textbf{Role} \\
\midrule
Qwen3.5-9B               & Qwen3.5 VL & Open-source  & 9\,B  & Default judge \\
Qwen2.5-VL-7B-Instruct   & Qwen2.5 VL & Open-source  & 7\,B  & Same family, earlier generation \\
GPT-5.4                  & GPT (OpenAI) & Closed-source & --- & Cross-family closed-source judge \\
\bottomrule
\end{tabular}
\end{table}

\subsubsection{Video generators}
We score four open-source text-to-video models that span $1.3\,\text{B}$--$14\,\text{B}$ parameters, two resolution tiers, and two output frame rates.
Every model is run on the full \phymetric{} suite of 1{,}115 scenes with the same caption as the video generation prompt.

\begin{table}[h]
\centering
\small
\setlength{\tabcolsep}{4pt}
\caption{Text-to-video generators evaluated by \phymetric{}.}
\label{tab:t2v_generators}
\begin{tabular}{llcccccc}
\toprule
\textbf{Generator} & \textbf{Architecture} & \textbf{Params} & \textbf{Resolution} & \textbf{Frames} & \textbf{FPS} & \textbf{Dur. (s)} & \textbf{Scenes} \\
\midrule
CogVideoX-5B    & DiT (T2V)      & 5\,B    & 720$\times$480  & 81  & 16 & 5.06 & 1{,}115 \\
Wan2.1-1.3B     & Wan T2V        & 1.3\,B  & 832$\times$480  & 81  & 16 & 5.06 & 1{,}115 \\
Wan2.2-14B      & Wan MoE T2V    & 14\,B   & 832$\times$480 & 81 & 16 & 5.06 & 1{,}115 \\
Cosmos-7B       & Cosmos Diffusion T2W & 7\,B & 1280$\times$704 & 121 & 30 & 4.03 & 1{,}115 \\
\bottomrule
\end{tabular}
\end{table}

\subsubsection{Evaluation cases}
\label{appen:evaluation_cases}
We illustrate the end-to-end evaluation pipeline on three representative scenes drawn from different physics domains.
For each case we show (i) the scene caption used verbatim as the T2V prompt, (ii) a compact rendering of the structured physical prior, (iii) one of the eight balanced binary questions, and (iv) the judge's parsed response.
The frame indices listed in each box are the benchmark indices passed to the VLM judge; the corresponding visual frames are sampled from the generated video.


\begin{evalcasebox}[title={\caseicon{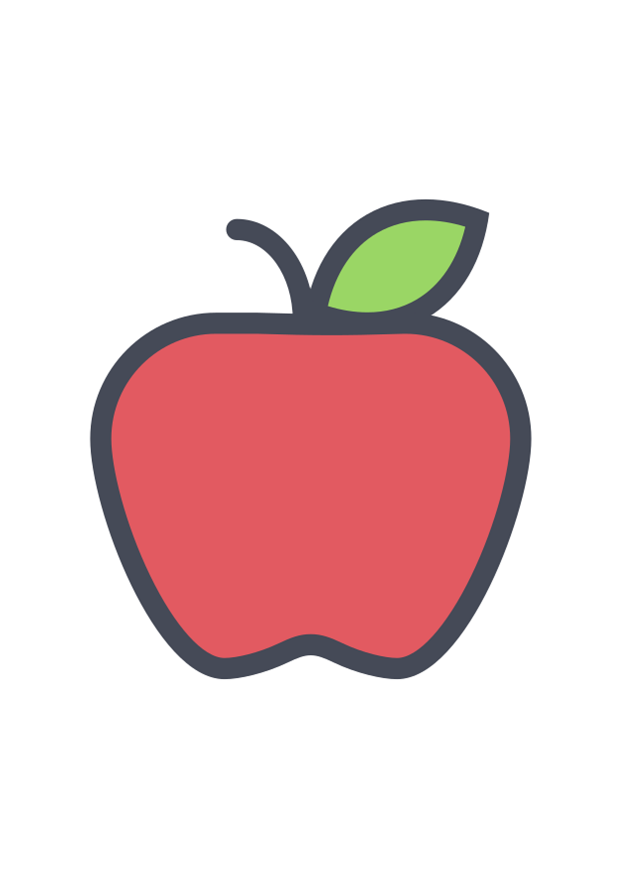}~Case 1 \textemdash{} Gravity\_Earth: ``Apple released from a tree branch''}]
\small
\textbf{Generator:} CogVideoX-5B \quad\\
\textbf{Frame indices shown to judge:} [20, 60, 100] / 120 \\[2pt]
\textbf{Scene caption:} ``A ripe apple is released from a tree branch above a grassy lawn and falls under Earth's gravity until it lands on the ground.'' \\[-4pt]
\evalframerow
  {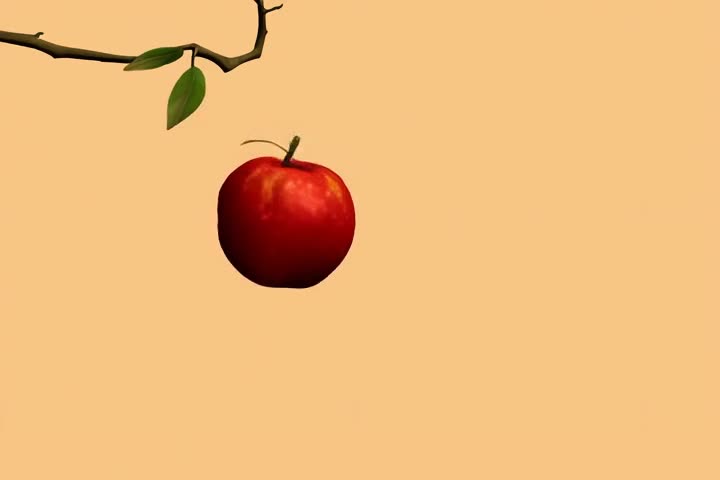}{frame 20 (\textasciitilde0.67\,s)}
  {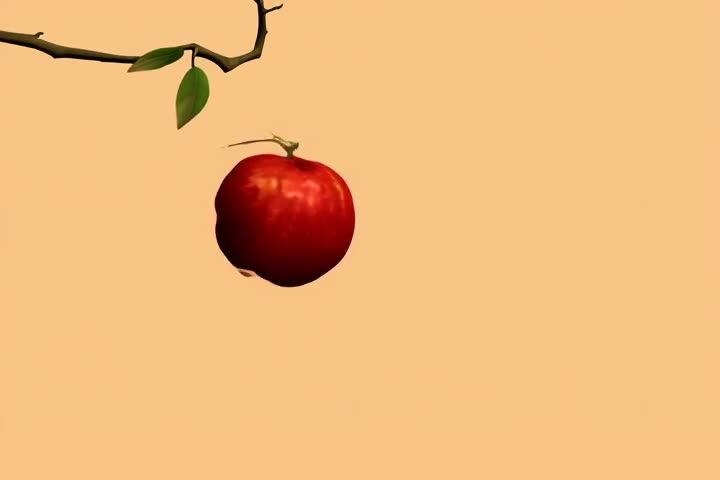}{frame 60 (\textasciitilde2.00\,s)}
  {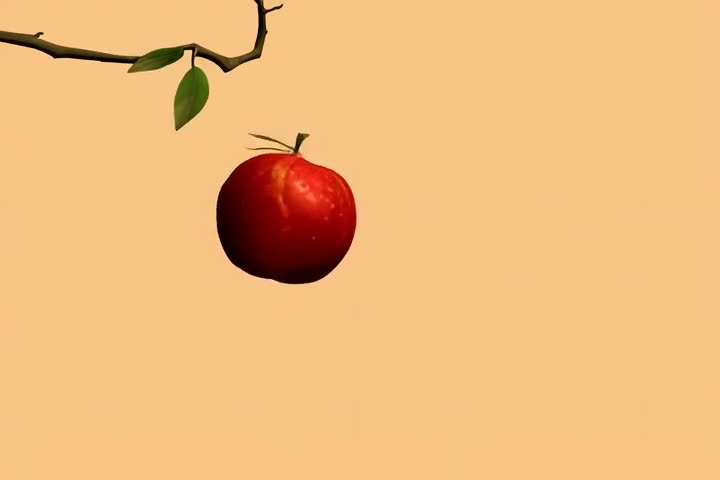}{frame 100 (\textasciitilde3.33\,s)}
\textit{\textbf{Structured physical prior (excerpt).}}
\begin{itemize}
  \item \textbf{Physical law:} Constant downward acceleration $g\!\approx\!9.8\,\text{m}/\text{s}^2$; vertical velocity grows linearly, distance as $\tfrac{1}{2}gt^2$.
  \item \textbf{Early:} apple just released, near the branch, vertical speed small.
  \item \textbf{Mid:} apple visibly accelerated, midway between branch and ground.
  \item \textbf{Late:} apple has reached the ground, no upward motion afterward.
  \item \textbf{Signature:} downward speed monotonically increases until contact; trajectory is straight-vertical, not floating, hovering, or rising.
\end{itemize}
\textit{\textbf{Question (\texttt{id=02}, dimension=motion, difficulty=easy, answer=1):}} \\
``Does the apple's downward speed visibly increase between the early and middle frames?'' \\[2pt]
\textit{\textbf{VLM judge response:}}\;\; \texttt{ANSWER: NO} \quad \texttt{CONFIDENCE: 0.83} \\
\texttt{REASON: The apple's vertical position is essentially unchanged across the three frames; it remains hovering just below the branch instead of accelerating downward toward the ground.} \\[2pt]
\textit{\textbf{Verdict:}} judge disagrees with ground truth (correct answer is YES) $\Rightarrow$ \textbf{question scored incorrect for CogVideoX-5B}.
This case illustrates a typical gravity failure: the generator produces a near-static apple instead of free fall, and the structured prior's ``downward speed monotonically increases'' signature lets the judge correctly flag the absence of acceleration.
\end{evalcasebox}

\begin{evalcasebox}[title={\caseicon{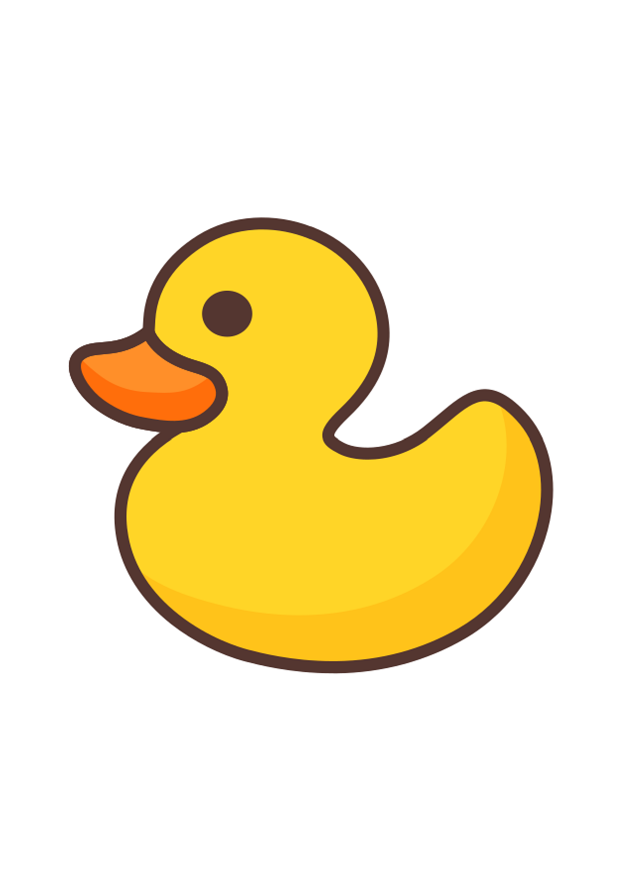}~~Case 2 \textemdash{} Buoyancy\_Float: ``Rubber duck placed in water''}]
\small
\textbf{Generator:} Wan2.1-1.3B \quad\\
\textbf{Frame indices shown to judge:} [25, 65, 105] / 120 \\[2pt]
\textbf{Scene caption:} ``A rubber duck is placed into water, displaces the water, and floats at the surface in a stable position.'' \\[-4pt]
\evalframerow
  {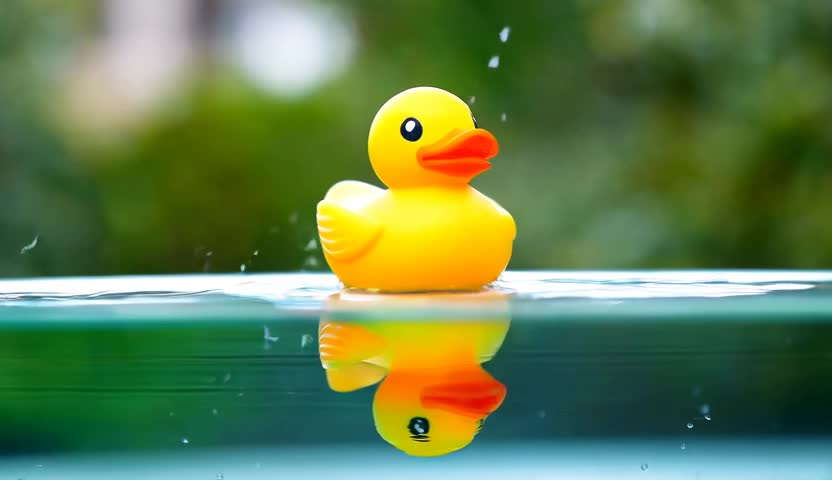}{frame 25 (\textasciitilde0.83\,s)}
  {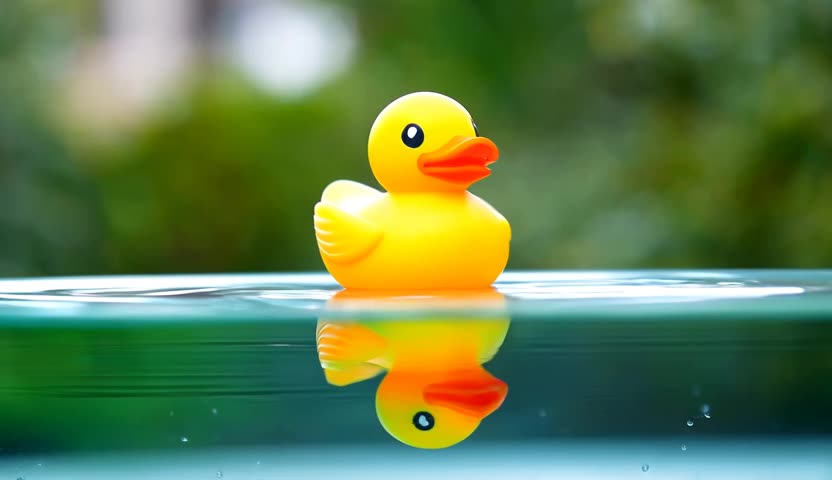}{frame 65 (\textasciitilde2.17\,s)}
  {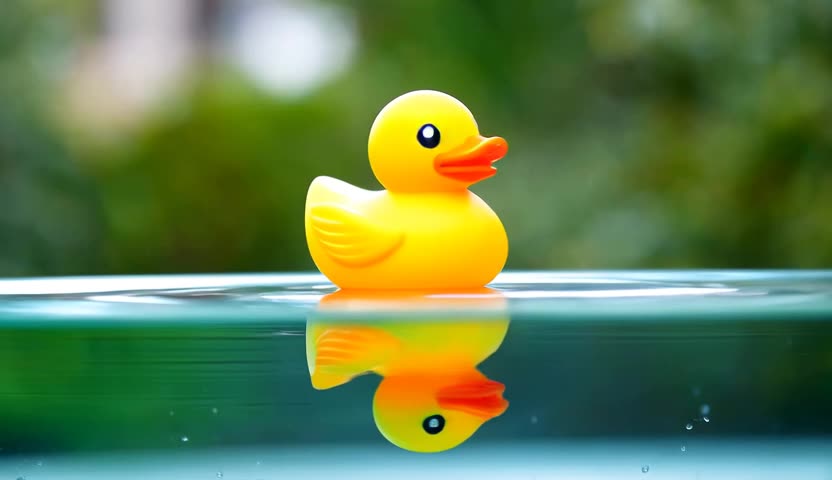}{frame 105 (\textasciitilde3.50\,s)}
\textit{\textbf{Structured physical prior (excerpt).}}
\begin{itemize}
  \item \textbf{Physical law:} Archimedes' principle; the duck's average density is below that of water, so the buoyant force balances gravity and the duck rests at the surface.
  \item \textbf{Early:} duck enters the water, briefly displacing fluid; small ripples and a short bobbing motion are visible.
  \item \textbf{Mid:} duck is partially submerged at the waterline, slowly settling into stable equilibrium.
  \item \textbf{Late:} duck floats steadily at the surface with its top half visibly above the water; no downward drift.
  \item \textbf{Signature:} the duck never fully submerges; the waterline cuts through its lower body for the entire clip.
\end{itemize}
\textit{\textbf{Question (\texttt{id=04}, dimension=spatial, difficulty=easy, answer=1):}} \\
``Does the rubber duck remain partly above the water surface across all the input frames?'' \\[2pt]
\textit{\textbf{VLM judge response:}}\;\; \texttt{ANSWER: YES} \quad \texttt{CONFIDENCE: 0.94} \\
\texttt{REASON: In every frame the duck's body sits clearly above the waterline with only its lower half submerged, consistent with stable floating.} \\[2pt]
\textit{\textbf{Verdict:}} judge agrees with ground truth $\Rightarrow$ \textbf{question scored correct}.
The mirror-image reflection the model renders below the waterline confirms the duck--surface contact relation that the prior's signature requires, allowing a high-confidence positive answer.
\end{evalcasebox}

\begin{evalcasebox}[title={\caseicon{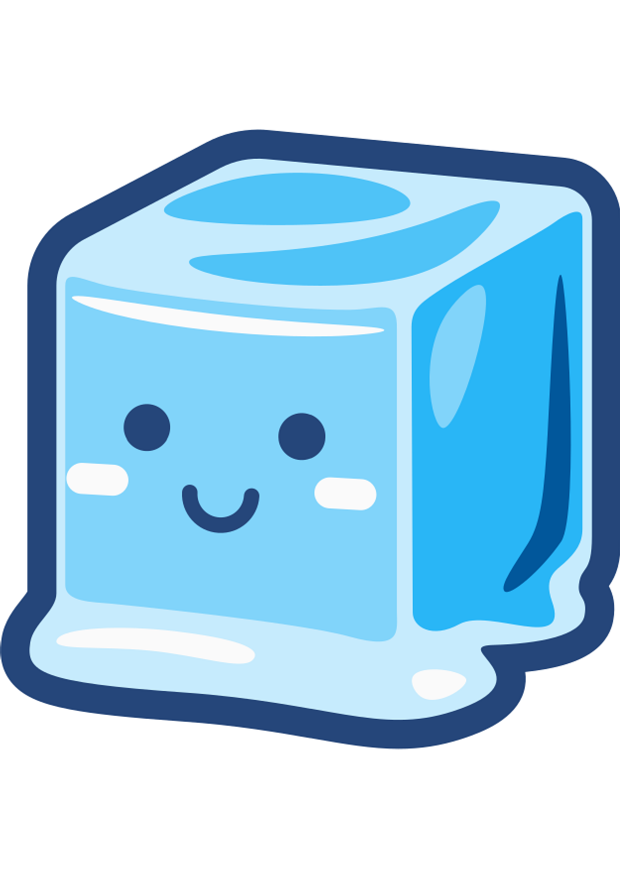}~Case 3 \textemdash{} Melting: ``Ice cube on a warm metal plate''}]
\small
\textbf{Generator:} Wan2.2-A14B \quad\\
\textbf{Frame indices shown to judge:} [15, 55, 100] / 120 \\[2pt]
\textbf{Scene caption:} ``An ice cube is placed on a warm metal plate at room temperature; over four seconds it gradually melts into a small puddle of clear water around its base.'' \\[-4pt]
\evalframerow
  {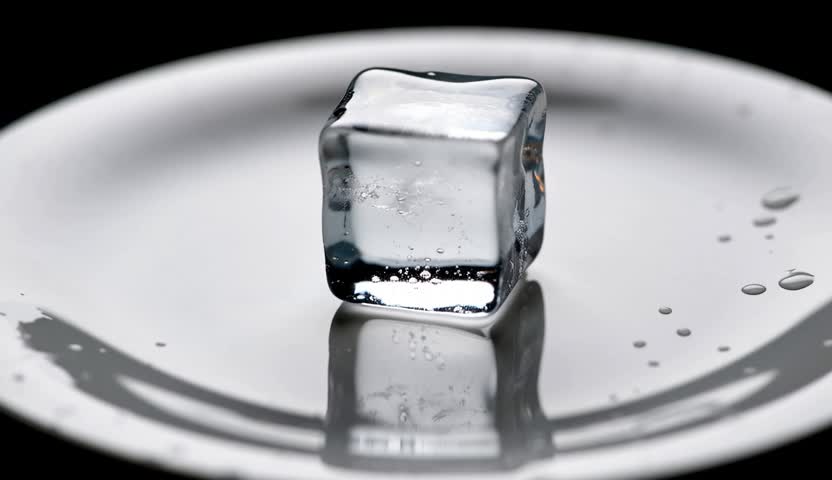}{frame 15 (\textasciitilde0.50\,s)}
  {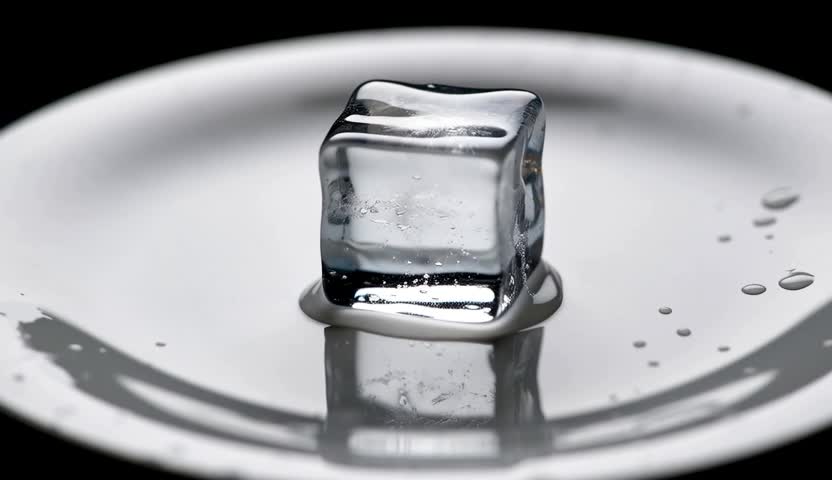}{frame 55 (\textasciitilde1.83\,s)}
  {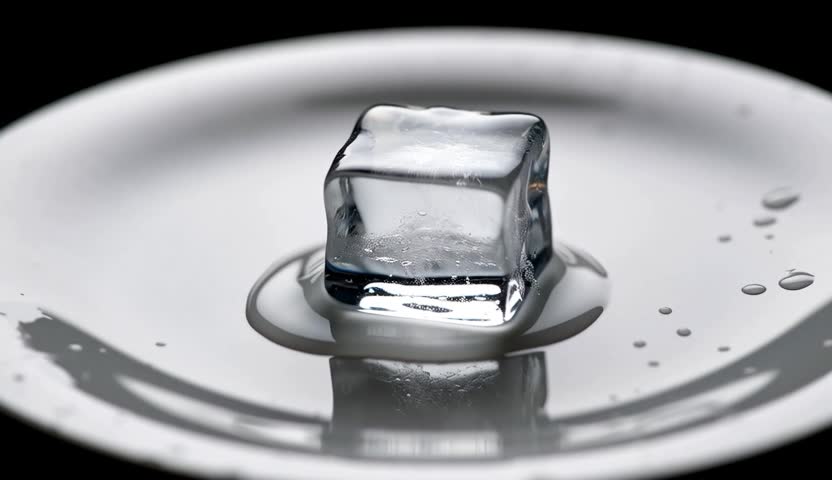}{frame 100 (\textasciitilde3.33\,s)}
\textit{\textbf{Structured physical prior (excerpt).}}
\begin{itemize}
  \item \textbf{Physical law:} Phase transition at melting point absorbs latent heat; solid volume shrinks while a liquid pool grows around the contact region.
  \item \textbf{Early:} ice cube intact, faint film of meltwater starting to form.
  \item \textbf{Mid:} cube visibly smaller, surrounding puddle larger and clearer.
  \item \textbf{Late:} only a small remnant of ice (or none) on top of a clear water pool.
  \item \textbf{Signature:} ice volume monotonically decreases; water volume monotonically increases; the cube does not regrow or stay constant.
\end{itemize}
\textit{\textbf{Question (\texttt{id=02}, dimension=shape, difficulty=hard, answer=1):}} \\
``Does the ice cube's volume visibly shrink between the early and late frames?'' \\[2pt]
\textit{\textbf{VLM judge response:}}\;\; \texttt{ANSWER: YES} \quad \texttt{CONFIDENCE: 0.84} \\
\texttt{REASON: The cube spans a noticeably smaller footprint on the plate in the late frame and is partly surrounded by a thin liquid film not present early on.} \\[2pt]
\textit{\textbf{Verdict:}} judge agrees with ground truth $\Rightarrow$ \textbf{question scored correct}.
The structured prior's signature (``ice volume monotonically decreases; water volume monotonically increases'') is what lets the judge convert the small visual change into a confident YES instead of treating it as ambiguous.
\end{evalcasebox}

\subsection{Experimental Details}
\label{appen:exp_details}

\subsubsection{Ablation setup}

\paragraph{\phymetric{} Component ablation.}
We isolate the two main components of \phymetric{} independently, (i) whether the structured physical prior is presented to the judge and (ii) whether the question is the scene-specific binary question or a single generic physical-plausibility question.
Crossing these two binary factors yields four conditions, summarized in Table~\ref{tab:partA}.
Comparing A1 against A2 isolates the contribution of the prior under fixed scene-specific questions; A2 against A3 isolates the contribution of scene-specific questions in the absence of a prior; and A1 against A4 tests whether scene-specific questions remain useful once the prior is provided.
All four conditions are run on every video model with all three VLM judges in Table~\ref{tab:vlm_judges} on a stratified subset of $N$ scenes per category.

\begin{table}[h]
\centering
\small
\setlength{\tabcolsep}{8pt}
\caption{Part A --- $2\times2$ component ablation: structured physical prior crossed with question specificity. A1 is the full system; A2--A4 each disable one component. All conditions are evaluated on every video model with all three VLM judges.}
\label{tab:partA}
\begin{tabular}{lccl}
\toprule
\textbf{Condition} & \textbf{Prior?} & \textbf{Specific Q?} & \textbf{Tests} \\
\midrule
A1 & \checkmark & \checkmark & Full system (reference) \\
A2 &            & \checkmark & Contribution of the structured prior \\
A3 &            &            & Raw VLM vision baseline \\
A4 & \checkmark &            & Prior without scene-specific QA \\
\bottomrule
\end{tabular}
\end{table}

\paragraph{Generic question prompt.}
Conditions A3 and A4 replace the scene-specific yes/no question with the fixed generic question shown below.
The generic question is intentionally minimal: it asks only whether the video respects the caption's physical behavior, with no reference to specific objects, frame indices, or expected outcomes, and therefore cannot benefit from the structured prior except indirectly through the unchanged scene caption.

\begin{tcolorbox}[
  enhanced, breakable, colback=red!4, colframe=red!40!black!88!,
  title=\textbf{Generic Question},
  fonttitle=\bfseries, left=4pt, right=4pt, top=2pt, bottom=2pt,
  boxrule=0.6pt, arc=2pt
]
\small\ttfamily
Does this video correctly simulate the physical behavior described in the scene caption? Answer YES or NO.
\end{tcolorbox}

\paragraph{Cross-source prior analysis.}
A natural concern is that improvements from the structured prior could be circular when the same model both authors the prior and judges the video.
To address this, Part B crosses three VLM judges with five prior sources -- a ``no prior'' baseline plus four priors authored by independent text LLMs (Qwen3.5-9B, Qwen2.5-VL-7B, DeepSeek-R1-Distill, and Mistral-7B), yielding $3\!\times\!5=15$ conditions per video model.
A judge that did not author the prior but still benefits from it (for example, the closed-source judge evaluated with a Qwen-authored prior) provides evidence that the prior encodes transferable physical content rather than judge-specific shortcuts.

\subsubsection{Human study}
\label{appen:human_study}
\paragraph{Scene subset.}
We hold out a stratified \textbf{100-scene subset} covering all 50 categories with $\sim$2 scenes per category, biased toward scenes whose physical signatures are unambiguous to a non-expert viewer.
Each scene is paired with its 8 expert-curated binary questions, giving \textbf{800 (scene, question) annotation units}.
We collect human responses against videos generated by all four T2V models.

\paragraph{Annotators}
We recruit \textbf{50 human annotators} from graduate-level or above academic groups, including students and researchers with general STEM backgrounds.
Annotators are not required to be physics experts, since the study is intended to measure whether the physical consequence is visually recognizable to an informed viewer.
Before annotation, each participant is given a short instruction page explaining the yes/no format, the meaning of sampled frames, and examples of valid and invalid judgments.
Annotators are asked to answer based only on the provided video evidence and scene caption, without using external resources.

\begin{figure}
    \centering
    \includegraphics[width=0.8\linewidth]{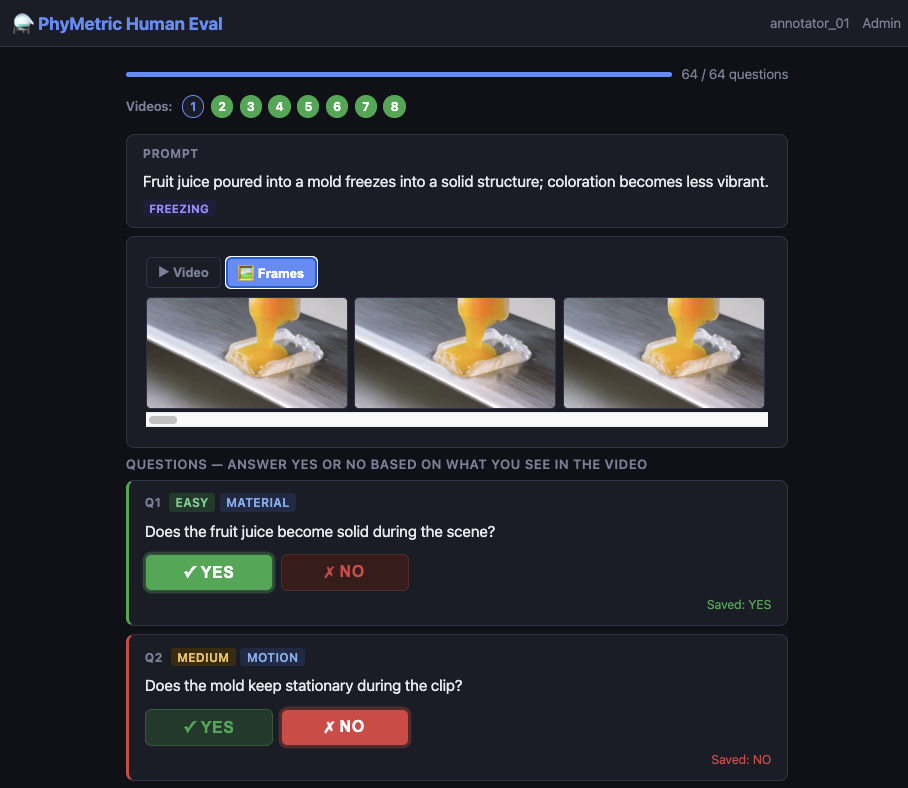}
    \caption{The interface of \phymetric{} human study.}
    \label{fig:human_eval_interface}
\end{figure}

\paragraph{Annotation interface.}
As shown in Figure~\ref{fig:human_eval_interface}, annotation is conducted through a self-hosted Flask web app that, for every (annotator, scene, question) triple, displays (i) the scene caption, (ii) the sampled frames at the question's frame indices, (iii) an optional embedded full-resolution video, and (iv) the binary question.
Question order is shuffled per annotator to control for ordering effects; the structured physical prior is \emph{not} shown to annotators, where they answer from the video alone.

\paragraph{Use of human responses.}
Aggregated human answers serve two purposes in the main paper.
First, they provide a \emph{ceiling} reference: per-question majority human accuracy upper-bounds what any automatic judge can achieve when the videos are visually ambiguous.
Second, they support the correlation with human judgment analysis (Section~\ref{sec:exp}), where per-scene human accuracy is correlated with the \phymetric{} score under each ablation condition to verify that improvements over baselines reflect alignment with human physical perception rather than artifacts of the prompt.

\subsection{Limitations and Future Work}
\label{appen:limitation}

\phymetric{} focuses on observable physical consequences in generated videos.
This design makes the benchmark interpretable and scalable, since each score can be traced to question-specific visual evidence.
However, it also limits what kinds of physics can be verified.
The current taxonomy covers 50 physical categories across six domains, but it is not exhaustive.
Some physical laws, such as energy conservation cannot be reliably judged from sparse sampled RGB frames alone.
Future versions can extend \phymetric{} with denser frame evidence, depth cues, object trajectories, tracking annotations, and simulation-derived states to support more quantitative physical checks.

Another limitation comes from the binary QA format.
Yes/no questions make the score easy to interpret and reduce ambiguity, but they simplify physical outcomes that may have multiple valid trajectories or degrees of correctness.
For example, a generated object may deform in a plausible but not exact way, or a fluid may spread with a reasonable but different temporal pattern.
Future work can combine binary QA with graded labels, multiple-choice physical outcomes, and continuous measurements to better capture partial correctness.

The automatic score also depends on the VLM judge.
Our ablations show that scene-specific questions are the most stable component, while structured priors affect different evaluators differently.
This suggests that future physical evaluation will benefit from stronger video-native VLMs, better uncertainty calibration, and selective human-in-the-loop verification, especially for subtle temporal or material changes.

Each model is evaluated with one generated video per scene for efficiency and controlled comparison across models.
This setting does not capture sampling variance across random seeds.
Future work should evaluate multiple generations per scene and report confidence intervals or bootstrap error bars.
Beyond evaluation, \phymetric{} can also support training: failed question labels, dimension-level scores, and structured priors can be used to construct reward models, preference pairs, or targeted fine-tuning data for physically grounded video generation.

Finally, \phymetric{} evaluates generated videos through question-aware sampled frames rather than full continuous video reasoning.
This design makes the evidence shown to the VLM explicit and reproducible, but it can also be rigid.
If the sampled frames miss a short-lived event, such as the exact moment of contact, deformation, splash, or phase transition, the VLM may judge from incomplete evidence.
Similarly, fixed frame indices may not align well with videos whose event timing differs from the expected temporal schedule.
Future work can replace fixed frame sampling with adaptive temporal localization, event-aware frame selection, or video-native evaluators that reason over the full clip while still reporting the evidence used for each decision.

\newpage
\input{checklist.tex}

\end{document}

%% file: sections/01introduction.tex
\section{Introduction}
\vspace{-0.2cm}
\label{sec:intro}

Text-to-video generation models~\cite{agarwal2025cosmos_arxiv2025,openai2024sora,wan2025wan_arxiv2025,yang2024cogvideox_iclr2025,veo3techreport,seedance2026seedance,miao2025advances,meng2026physical_acmmm2026,wang2026rigi} have achieved strong visual fidelity and are increasingly applied to dynamic-scene simulation, not only to clip-level visual generation~\cite{bruce2024genie_icml2024,meng2024towards_icml2025,balazadeh2025physics_iccv2025}.
In this regime, a generated video should not only look realistic frame by frame, but also evolve according to the physical and causal regularities implied by the prompt and scene.
For example, a falling object should accelerate downward, a sliced object should separate into pieces, and a buoyant object should remain supported by fluid rather than drift arbitrarily.
Conventional visual quality metrics~\cite{wang2004image_tip2004,unterthiner2019fvd,wang2024internvid_iclr2024}, however, operate largely at the frame level and cannot capture such temporal causality.
As a result, they may assign high scores to videos that look plausible at every frame but violate basic physical expectations~\cite{meng2025grounding_ijcai2025,kang2024far_icml2025}.

Recent benchmarks have started to address this mismatch by extending evaluation from surface-level realism toward world-aware and physics-oriented analysis~\cite{huang2024vbench_cvpr2024,liu2024evalcrafter_cvpr2024,zheng2025vbench_pami2025,wang2025videoverse_arxiv2025,meng2024towards_icml2025,weicophybench_2025,motamed2025travl_arxiv2025,meng2026phymagic_eccv2026}.
Despite this progress, many of them still rely on coarse plausibility assessments that ask whether a video looks physically reasonable as a whole, rather than verifying whether the key physical event implied by the prompt is actually realized.
These assessments aggregate diverse physical consequences into a single score, which can hide a missing or incorrect physical event despite a high overall score. 
Conversely, a low score reveals little diagnostic insight into which particular physical law was violated.
The result is a persistent gap between surface realism and verifiable physical plausibility.

\begin{figure}[t]
    \centering
    \includegraphics[width=0.86\linewidth]{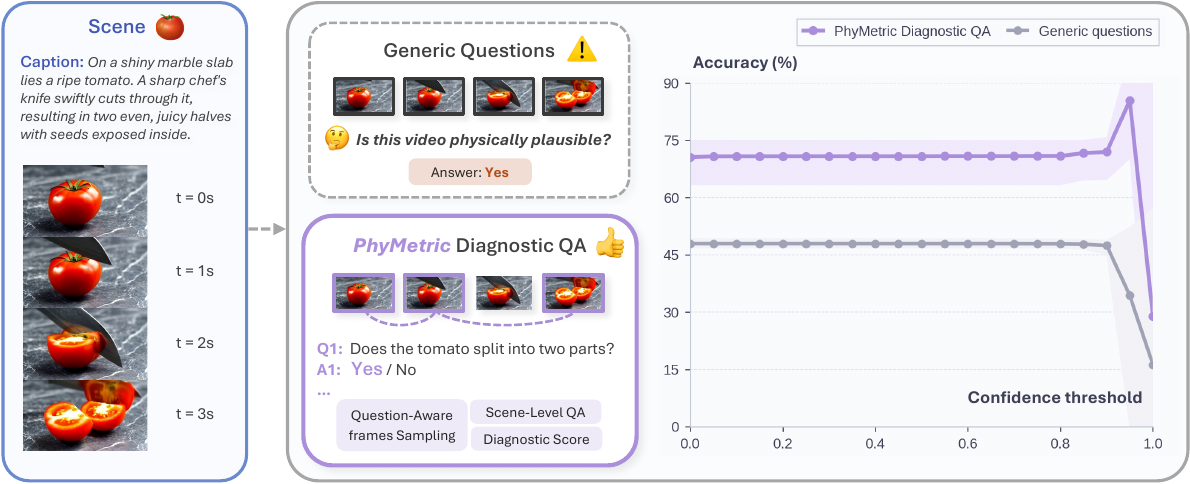}
    \caption{\small\textbf{Why holistic plausibility is insufficient for physical evaluation.}
    Generic plausibility questions tend to yield high-confidence but unreliable judgments by overlooking scene-specific physical consequences.
    \phymetric{} instead decomposes evaluation into structured diagnostic QA with question-aware frame sampling, enabling more robust and interpretable physical assessment.
    In the right subfigure, as confidence thresholds increase, generic-question accuracy degrades sharply and only achieves around 50\% accuracy, whereas structured scene-specific questions remain substantially more stable.
    }
    \label{fig:teaser_overview}
    \vspace{-12pt}
\end{figure}

We attribute this gap to two core challenges: \textbf{identifiability} and \textbf{expectation grounding}.
\emph{Identifiability} requires the benchmark to make explicit which physical consequence is being tested.
A single holistic prompt is too weak a probe for physical reasoning: it obscures which consequence is being tested and remains highly susceptible to linguistic drift or evaluator bias. 
As shown in Figure~\ref{fig:teaser_overview}, this lack of granularity can cause VLM-judge performance close to chance levels, even under high-confidence predictions.
Beyond identifiability, \emph{expectation grounding} requires the evaluator to know what physical behavior the scene implies before judging.
For instance, elastic and inelastic collisions both involve contact but yield fundamentally different correct outcomes.
Without a scene-specific prior, judges fall back on generic heuristics and often confuse semantic mismatches with genuine physical failures~\cite{wang2025videoverse_arxiv2025,meng2024towards_icml2025,zhang2026physion_arxiv2025}.

Motivated by these challenges, we introduce \phymetric{}, a scene-level diagnostic benchmark that replaces holistic plausibility scoring with scene-specific balanced binary QA.
For each physical scene, \phymetric{} issues multiple yes/no questions, each targeting a verifiable physical consequence that can be checked from a small set of question-aware frames.
This design directly addresses identifiability: it makes the tested consequence explicit and keeps the chance baseline interpretable by construction.
To address expectation grounding, every scene is further paired with a structured physical prior that summarizes the governing law, the expected temporal sequence, the observable signatures, and common failures, anchoring the VLM evaluator in the correct physical context before video inspection.
\phymetric{} thus supports a focused evaluative claim, namely whether a generated video exhibits the specific physical consequences implied by the scene, while remaining agnostic to broader questions of latent physical understanding or generalization beyond observed dynamics.

In total, \phymetric{} comprises 1{,}115 scenes from 50 physical categories across six domains: mechanics, fluids, materials, thermal processes, electromagnetism, and chemistry (Figure~\ref{fig:taxonomy}).
These categories are chosen as a coverage map over the dynamics that recur in generated video, rather than as a topical list, so that \phymetric{} probes a diverse range of physical behaviors instead of overfitting to familiar motions such as falling or bouncing.
Each scene is annotated with eight expert-verified binary questions, yielding 8{,}920 QA pairs spanning 12 evaluation dimensions.
At evaluation time, \phymetric{} runs a three-step pipeline that first performs question-aware frame sampling around the queried event, then conditions the VLM evaluator on the structured physical prior, and finally aggregates binary responses hierarchically into scene-, dimension-, category-, and model-level diagnostic scores.
Across four open-source text-to-video models and three VLM judges, we find that scene-specific QA is what enables non-trivial evaluator accuracy: replacing it with a generic plausibility query collapses all three judges to chance, while structured priors provide a complementary, evaluator-dependent signal that improves human alignment of the resulting scores.

In summary, our main contributions are as follows:
\begin{itemize}
\item We introduce \phymetric{}, a scene-level diagnostic benchmark for physical plausibility in text-to-video generation, comprising 1{,}115 scenes over 50 categories and six physical domains, paired with 8{,}920 answer-balanced binary questions that span 12 evaluation dimensions.
\vspace{-0.05cm}
\item We recast physical plausibility evaluation from holistic scoring to per-scene balanced binary QA grounded in a structured physical prior, and show empirically that scene-specific questions are necessary for non-trivial evaluator accuracy while the prior provides a complementary, evaluator-dependent signal that improves human alignment.
\vspace{-0.05cm}
\item We validate \phymetric{} across four open-source T2V models and three VLM judges, demonstrating consistent model rankings across VLM judges, distinctive dimension-level capability profiles, and systematic weaknesses in current systems that are not visible to holistic scoring.
\end{itemize}

%% file: sections/02relatedWorks.tex
\section{Related Works}
\vspace{-0.2cm}
\label{sec:related}

\noindent\textbf{General Video Generation Benchmarks.}
Video generation evaluation has moved from single-score metrics, such as FVD~\cite{unterthiner2019fvd} and CLIP-based similarity~\cite{wang2024internvid_iclr2024}, toward multi-dimensional benchmark suites.
VBench~\cite{huang2024vbench_cvpr2024} addresses this limitation by decomposing video generation quality into a hierarchy of fine-grained dimensions covering quality and semantics, while EvalCrafter~\cite{liu2024evalcrafter_cvpr2024}, FETV~\cite{liu2023fetv_neurips2025}, and T2V-CompBench~\cite{sun2025t2v_cvpr2025} provide complementary multi-dimensional and compositional protocols.
More recent benchmarks reach further than surface-level visual quality and probe whether generated videos satisfy richer notions of real-world validity, often by reformulating evaluation as a human- or model-aligned reasoning problem.
Video-Bench~\cite{han2025video_cvpr2025} formulates video generation evaluation as a human-aligned MLLM-based scoring problem with chain-of-query reasoning, and ETVA~\cite{guan2025etva_iccv2025} further pushes this direction by generating dense, prompt-derived questions to probe text-video alignment at fine granularity.
Building on these efforts, VBench2.0~\cite{zheng2025vbench_pami2025} formalises the goal as intrinsic faithfulness covering human fidelity, controllability, creativity, physics, and commonsense, and VideoVerse~\cite{wang2025videoverse_arxiv2025} introduces hidden semantics to test whether T2V models can infer unstated event consequences rather than merely follow explicit prompt content.
Across these benchmarks, physics typically appears as one component of a larger quality rubric, leaving open how scene-specific physical consequences should be verified for individual generations.

\noindent\textbf{Physics-Oriented Video Generation Benchmarks.}
A growing line of work focuses directly on physical commonsense and physical laws.
VideoPhy~\cite{bansal2024videophy_iclr2025} and VideoPhy-2~\cite{bansal2025videophy_arxiv2025} evaluate semantic adherence and physical commonsense across solids, fluids, and multi-material interactions.
PhyGenBench~\cite{meng2024towards_icml2025} constructs prompts around explicit physical laws and introduces PhyGenEval for hierarchical physical evaluation.
Other works move beyond direct plausibility scoring by anchoring evaluation in alternative forms of physical evidence.
Morpheus~\cite{zhang2025morpheus_arxiv2025} grounds evaluation in real physical experiments and physical invariants.
LikePhys~\cite{yuan2025likephy_arxiv} avoids direct VLM judging and instead measures intuitive physics understanding through likelihood preference between physically valid and invalid samples, while Physion-Eval~\cite{zhang2026physion_arxiv2025} provides expert human reasoning traces for diagnosing temporally localized physical glitches.
Evaluator-focused works also show the difficulty of physical judgment.
PhysBench~\cite{chow2025physbench_iclr2025} shows that broad VLM judges miss physically critical details when asked a single broad question and recover only when paired with explicit physical priors.
CoPhyBench~\cite{weicophybench_2025} improves diagnosticity by decomposing physical reasoning into prediction, calculation, and counterfactual inference,
and TRAVL~\cite{motamed2025travl_arxiv2025} shows that current VLM evaluators struggle with implausibility detection and proposes trajectory-aware training to improve their behavior.

\phymetric{} addresses these recurring limitations through target design decisions.
While prior works often rely on holistic scoring that obscures specific failure modes, \phymetric{} replaces broad assessments with balanced binary questions to ensure \emph{identifiability}, allowing every score to be traced back to concrete physical checks.
To solve the problem of expectation grounding, where evaluators lack the context to distinguish physical errors from semantic ones, we pair each scene with a structured physical prior that anchors the judge in the governing law before inspection. 
Furthermore, as shown in Table~\ref{tab:benchmark_compare}, \phymetric{} implements balanced labels to provide an interpretable 50\% chance baseline, and utilizes question-aware frame sampling to link judgments to time-sensitive physical evidence. 
Compared with QA-based benchmarks like ETVA~\cite{guan2025etva_iccv2025}, \phymetric{} anchors its queries to objective physical consequences rather than text-alignment coverage, bridging the gap between surface-level realism and verifiable physical plausibility.


\begin{table*}[t]
\centering
\small
\caption{
\small
\textbf{Comparison with existing video generation benchmarks.}
We group prior benchmarks into two main aspects.
\cmark denotes explicit support, \xmark denotes no support, and \pmark denotes partial support.
\textbf{Balanced Labels} indicates whether positive/negative answers are balanced for an interpretable chance baseline.
Compared with broad video-generation and physics-oriented benchmarks, \phymetric{} combines large-scale scene coverage, scene-specific yes/no QA, structured physical priors, and balanced labels for physical plausibility evaluation.
}
\label{tab:benchmark_compare}
\renewcommand{\arraystretch}{1.05}
\footnotesize
\resizebox{\textwidth}{!}{%
\begin{tabular}{llcccccc}
\toprule
\textbf{Benchmark} 
& \textbf{Main target} 
& \textbf{Scenes / Prompts} 
& \textbf{Questions} 
& \textbf{QA Format} 
& \textbf{Physics / Prior} 
& \textbf{Balanced Labels} 
& \textbf{Year} \\
\midrule
\multicolumn{8}{l}{\textit{General and fine-grained video generation evaluation}} \\
\midrule
VBench~\cite{huang2024vbench_cvpr2024} 
& General video quality 
& -- 
& -- 
& \xmark 
& \pmark 
& \xmark 
& 2024 \\
EvalCrafter~\cite{liu2024evalcrafter_cvpr2024} 
& General video quality 
& 700 
& -- 
& \xmark 
& \xmark 
& \xmark 
& 2024 \\
FETV~\cite{liu2023fetv_neurips2025} 
& Fine-grained T2V eval. 
& 619 
& -- 
& \xmark 
& \xmark 
& \xmark 
& 2025 \\
T2V-CompBench~\cite{sun2025t2v_cvpr2025} 
& Compositional T2V 
& 700 
& -- 
& \pmark 
& \xmark 
& \xmark 
& 2025 \\
Video-Bench~\cite{han2025video_cvpr2025} 
& Human-aligned gen. eval. 
& 419 
& -- 
& \pmark 
& \pmark 
& \xmark 
& 2025 \\
VBench-2.0~\cite{zheng2025vbench_pami2025} 
& Intrinsic faithfulness 
& -- 
& -- 
& \pmark 
& \pmark 
& \xmark 
& 2025 \\
ETVA~\cite{guan2025etva_iccv2025} 
& Text-video alignment 
& 2{,}000 
& 12K 
& \cmark 
& \pmark 
& \xmark 
& 2025 \\
\midrule
\multicolumn{8}{l}{\textit{Temporal, world-model, and physics-oriented evaluation}} \\
\midrule
ChronoMagic-Bench~\cite{yuan2024chronomagic_neurips2024} 
& Time-lapse transformation 
& 1{,}649 
& -- 
& \xmark 
& \pmark 
& \xmark 
& 2024 \\
WorldScore~\cite{duan2025worldscore_iccv2025} 
& World generation 
& 3{,}000 
& -- 
& \xmark 
& \pmark 
& \xmark 
& 2025 \\
VideoVerse~\cite{wang2025videoverse_arxiv2025} 
& World-model capability 
& 300 
& 793 
& \cmark 
& \pmark 
& \xmark 
& 2025 \\
PhyGenBench~\cite{meng2024towards_icml2025} 
& Physical laws 
& 160 
& -- 
& \pmark 
& \cmark 
& \xmark 
& 2025 \\
VideoPhy~\cite{bansal2024videophy_iclr2025} 
& Physical commonsense 
& 688 
& -- 
& \xmark 
& \pmark 
& \xmark 
& 2025 \\
VideoPhy-2~\cite{bansal2025videophy_arxiv2025} 
& Physical commonsense 
& 688 
& -- 
& \xmark 
& \pmark 
& \xmark 
& 2025 \\
Morpheus~\cite{zhang2025morpheus_arxiv2025} 
& Physical invariants 
& 80 
& -- 
& \xmark 
& \cmark 
& \xmark 
& 2025 \\
LikePhys~\cite{yuan2025likephy_arxiv} 
& Likelihood preference 
& -- 
& -- 
& \xmark 
& \cmark 
& \pmark 
& 2025 \\
Physion-Eval~\cite{zhang2026physion_arxiv2025} 
& Human glitch diagnosis 
& -- 
& -- 
& \xmark 
& \pmark 
& \xmark 
& 2025 \\
\midrule
\rowcolor{gray!10}
\textbf{\phymetric{} (ours)} 
& \textbf{Physical plausibility} 
& \textbf{1{,}115} 
& \textbf{8{,}920} 
& \textbf{\cmark} 
& \textbf{\cmark} 
& \textbf{\cmark} 
& 2026 \\
\bottomrule
\end{tabular}%
}
\vspace{-5pt}
\end{table*}

%% file: sections/03benchmark.tex
\section{The \phymetric{} Benchmark}
\vspace{-0.2cm}
\label{sec:phymetric}

\phymetric{} addresses the two challenges in Section~\ref{sec:intro} with two design decisions that jointly guide dataset curation and evaluation.
This section motivates each decision and describes its impact on both benchmark design and evaluation protocol (Sections~\ref{subsec:identifiability} and~\ref{subsec:locality}), before consolidating the resulting benchmark statistics and release artifacts.


\subsection{Identifiability through scene-specific binary QA}
\vspace{-0.2cm}
\label{subsec:identifiability}
A central failure mode of holistic physical evaluation is lack of identifiability.
When a judge is asked a broad question such as \textit{``Is this video physically plausible?''}, a low score does not reveal which physical condition failed.
The answer may be affected by visual quality, object appearance, prompt mismatch, or the judge's prior knowledge, even when the target physical event is only one part of the clip.
This makes the resulting score hard to interpret and hard to use for model diagnosis.

\phymetric{} addresses this problem by formulating physical evaluation as scene-specific binary question answering.
For each scene $s_i$, we define a set of observable questions:
\vspace{-0.2em}
\begin{equation}
    \small\mathcal{Q}_i = \{(q_{i,j}, y_{i,j}, d_{i,j})\}_{j=1}^{M},
\end{equation}
where $q_{i,j}$ is a binary yes/no question, $y_{i,j} \in \{\texttt{yes}, \texttt{no}\}$ is the ground truth answer, and $d_{i,j}$ denotes the evaluation dimension associated with the question.
In our benchmark, each scene contains $M=8$ binary questions.
These questions are not generic templates shared across all scenes.
They are written for the concrete objects, environments, actions, and expected physical consequences of the scene.
As illustrated in Figure~\ref{fig:teaser_overview}, instead of asking whether a video of cutting a tomato is physically plausible, \phymetric{} asks whether the knife contacts the tomato, whether the tomato separates after the cut, and whether the two pieces remain visually consistent.
We provide the whole scene construction in Appendix~\ref{app:category_cards} and the examples of scene-specific questions in Appendix~\ref{appen:evaluation_cases}.

This design turns a vague physical judgment into a set of identifiable checks.
Each question corresponds to a specific hypothesis about the generated video.
If a model receives a low scene score, the failed questions reveal whether the error comes from motion direction, object interaction, material response, or another physical dimension.
Thus, the benchmark score is not only a ranking signal, but also a diagnostic signal for physical plausibility.
The binary format also improves score calibration, with each response mapped to a binary truth value, the resulting question score is defined as:
\vspace{-0.1em}
\begin{equation}
    \small a_{i,j}(V) =
    \mathbbm{1}\big[\hat{y}_{i,j}(V) = y_{i,j}\big],
\end{equation}
where $\hat{y}_{i,j}(V)$ is the predicted answer for generated video $V$.
The scene-level score is then the average question accuracy:
\begin{equation}
    \small S_i(V) =
    \frac{1}{M}
    \sum_{j=1}^{M}
    a_{i,j}(V).
\end{equation}
By averaging results across scenes, we establish a robust model-level metric. 
The balanced nature of our binary labels provides a natural $50\%$ baseline, effectively reducing the influence of random guessing and answer bias. 
We emphasize that the transition to a binary QA format is not a reduction of physical complexity, but a prerequisite for rigorous verification. 
Many physical failures in generated videos are local and conditional.
A model may satisfy one consequence of a scene while failing another.
By using multiple targeted questions per scene, \phymetric{} captures this mixed behavior while keeping evaluation automatic and comparable across models.

\subsection{Expectation grounding through video-independent priors}
\vspace{-0.2cm}
\label{subsec:locality}

A scene-specific question is necessary but not sufficient.
A VLM judge that has never been told what the scene is supposed to do will answer the question using whichever heuristics the visual evidence happens to invoke, and an evaluator that receives the whole video without temporal guidance may attend to frames where the queried event has not yet occurred~\cite{chen2026t2vworldbench_cvpr2026}.
Expectation grounding therefore requires the evaluator to know two things before it judges: \emph{what} physical behavior the scene implies, and \emph{when} in the video to look for it.
\phymetric{} addresses both through a video-independent structured prior $K_s$.

\paragraph{What to expect.}
For each scene $s_i$, we construct a structured prior
\begin{equation}
   \small K_{s_i} = \{\ell_i, o_i, m_i, e_i, r_i, h_i\},
\end{equation}
where $\ell_i$ denotes the governing physical law, $o_i$ the involved objects, $m_i$ the material or state information, $e_i$ the expected event sequence, $r_i$ likely failure modes, and $h_i$ dimension-level evaluation hints.
For example, in a tomato-slicing scene, $\ell_i$ describes cutting as blade-induced shear failure, $o_i$ includes the knife and tomato, $m_i$ specifies a soft deformable tomato, $e_i$ expects contact followed by separation into two halves, $r_i$ lists failures such as no cut or floating pieces, and $h_i$ tells the judge to check shape, interaction, and appearance consistency.
$K_{s_i}$ is constructed from the scene caption and category metadata only, and is independent of the generated video $V$.
At evaluation time, the four core fields $\{\ell_i, e_i, r_i, h_i\}$ are passed to the VLM judge as textual context, while $\{o_i, m_i\}$ are auxiliary metadata used during scene and question construction (full schema in Appendix~\ref{appen:priors_construction}).
This independence is a deliberate choice: a prior that conditions on $V$ would adapt to the same visual heuristics the evaluator is being asked to bypass.

\paragraph{When to look.}
Physical events are temporally local: a question about initial contact should focus on the moment two objects meet, while a question about final state should focus on the outcome after the interaction.
Each question $q_{i,j}$ is therefore associated with one of five temporal anchors recorded in $h_i$ (initial, interaction, transition, final, or full sequence), which marks where the relevant evidence is expected to appear in the reference clip.
For a generated video $V$ with $N_V$ frames, anchor positions from a reference clip with $N_{\mathrm{ref}}$ frames are mapped to $V$ by
\begin{equation}
   \small f' = \mathrm{round}\!\left(\frac{f}{N_{\mathrm{ref}} - 1}(N_V - 1)\right),
\end{equation}
and the target frame together with its neighbors forms the evidence set $\mathcal{F}_{i,j}(V)$ passed to the judge.
This linear mapping assumes that the temporal location of an event in $V$ is approximately proportional to its location in the reference clip.
This assumption holds when the model preserves the prompted event ordering, but it may fail under conditions of temporal compression or reordering.
We further discuss this limitation in Appendix~\ref{appen:limitation}.

\paragraph{Joint role.}
The structured prior $K_s$ and the question-aware frame set $\mathcal{F}_{i,j}(V)$ are complementary: $K_s$ tells the evaluator what physical process to expect, and $\mathcal{F}_{i,j}(V)$ tells it where in the video to verify that expectation.
Neither alone is sufficient: a prior without temporally localized evidence describes a behavior the evaluator cannot anchor visually, and localized frames without a prior leave the evaluator to guess which physical consequence is being tested.

\begin{figure}[t]
    \centering
    \includegraphics[width=0.99\linewidth]{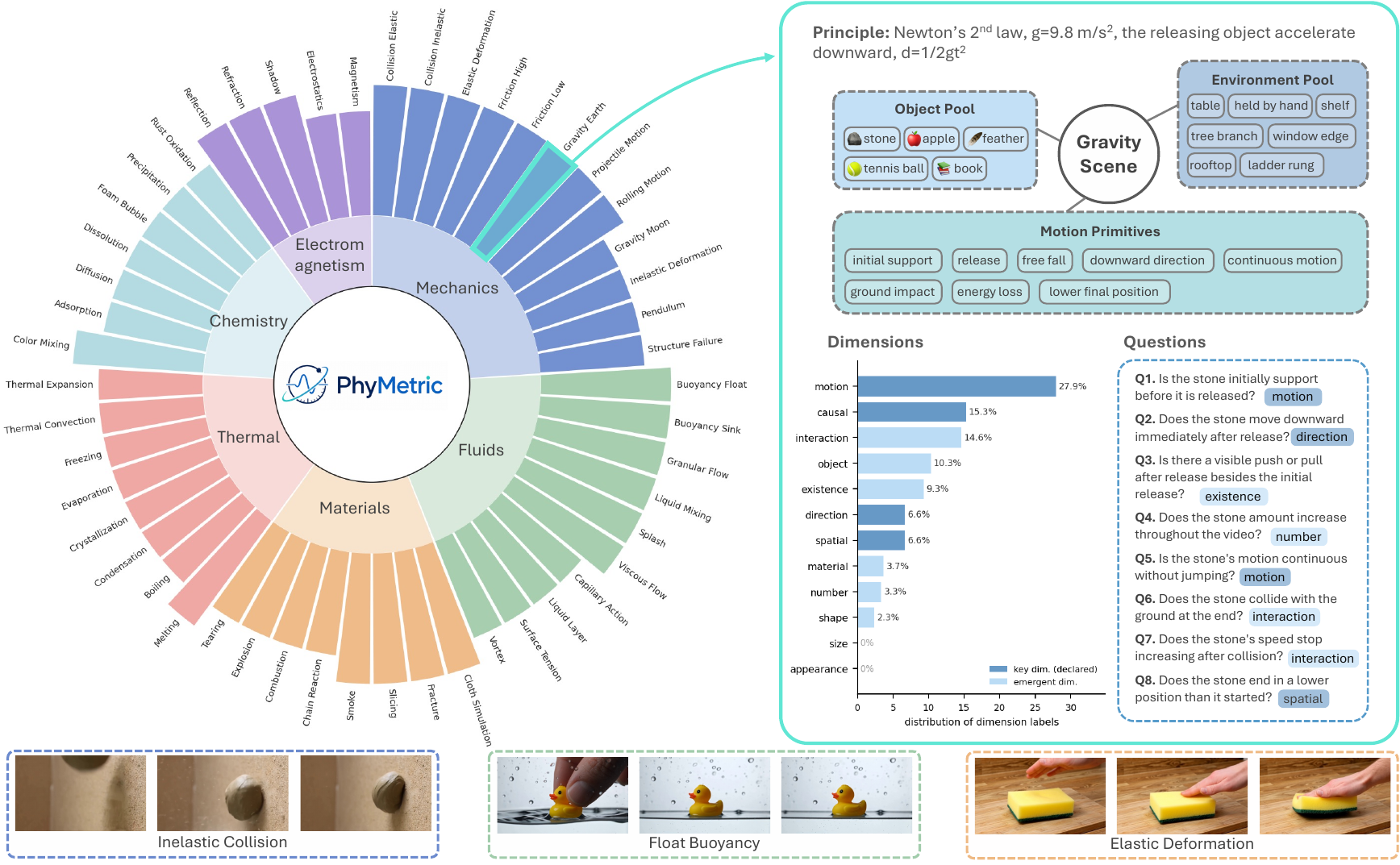}
    \caption{\small\textbf{\phymetric{} benchmark construction.}
    \phymetric{} organizes physical video generation into a structured taxonomy with 50 categories across six domains.
    Each category is defined by physical principles, object pools, environment pools, and diagnostic dimensions.
    These metadata are expanded into diverse scenes and balanced scene-specific binary questions.
    The bottom examples show representative physical events.}
    \label{fig:taxonomy}
    \vspace{-5pt}
\end{figure}

\noindent\textbf{Benchmark summary and release}
\label{subsec:benchmark_summary.}
As shown in Figure~\ref{fig:taxonomy}, \phymetric{} contains 1,115 physical scenes covering 50 fine-grained physical categories across six physical domains.
Each scene is paired with eight binary questions, resulting in 8,920 scene-specific QA pairs.
The questions span 12 evaluation dimensions, covering both low-level visual evidence and higher-level physical consequences.
The benchmark also provides structured physical priors for all scenes, including the expected event sequence, relevant physical principle, dimension hints, and common failure modes.
These components jointly support the evaluation pipeline illustrated in Figure~\ref{fig:evaluation_pipeline}, where question-aware frame sampling and structured priors transform each generated video into interpretable scene-level physical diagnosis.
Compared with holistic video-level scoring, \phymetric{} provides three key properties.
First, it is identifiable: each score can be traced back to concrete question-level physical checks.
Second, it is grounded: the expected behavior is defined before observing generated videos.
Third, it is evidence-localized: question-aware frame sampling links each judgment to the relevant temporal region of the video.

%% file: sections/05experiment.tex
\section{Experiments}
\vspace{-0.2cm}
\label{sec:exp}


Our experiments address three questions: whether \phymetric{} separates current T2V models at scale, which physical dimensions remain difficult, and whether scene-specific QA and structured physical priors are necessary for reliable evaluation.
Accordingly, we organize the experiments into three parts.
We first report the main model comparison on the full benchmark and a controlled 100-scene subset.
We then analyze model performance across 12 evaluation dimensions to identify systematic physical weaknesses.
Finally, we conduct ablation studies and human validation to test the role of scene-specific questions, structured physical priors, and our robustness on VLM judges.

\subsection{Experimental Setup}
\vspace{-0.2cm}
\label{subsec:experimentalSetup}

We evaluate four open-source text-to-video models: CogVideoX-5B~\cite{yang2024cogvideox_iclr2025}, Wan2.1-1.3B~\cite{wan2025wan_arxiv2025}, Wan2.2-14B~\cite{wan2025wan_arxiv2025}, and Cosmos-7B~\cite{agarwal2025cosmos_arxiv2025}.
Each model generates one video for each scene using the original scene caption as input, without model-specific prompt rewriting.
The main evaluation is conducted on the full \phymetric{} benchmark with 1{,}115 scenes from 50 physical categories.
For human validation and ablation studies, we use a controlled 100-scene subset, sampled as two scenes per category, which preserves category-level coverage while reducing annotation and multi-evaluator cost.

We use Qwen3.5-9B~\cite{yang2025qwen3} as the default VLM judge, and additionally evaluate with Qwen2.5-VL-7B-Instruct~\cite{bai2025qwen2} and GPT-5.4~\cite{singh2025openai_gpt5} to test evaluator robustness.
Each judge answers the scene-specific yes/no questions from question-aware sampled frames, the scene caption, and the structured physical prior $K_s$.
We report \emph{Overall} accuracy and \emph{Weighted} accuracy, where Overall is the mean binary correctness over all evaluated questions and Weighted incorporates the VLM judge's self-reported confidence.
Implementation details are provided in Appendix~\ref{appen:scene_construction} and~\ref{appen:priors_construction}.

For the human study, 50 annotators answer the same yes/no questions on the controlled 100-scene subset.
We use human responses to measure whether automatic scores align with human-perceived physical plausibility, rather than to replace benchmark labels.
We report scene-level Spearman correlation $\rho_{\text{human}}$ between VLM scores and human scores.
More details about the annotation interface, quality control, and agreement analysis are provided in Appendix~\ref{appen:human_study}.

\begin{figure}[t]
    \centering
    \includegraphics[width=0.96\linewidth]{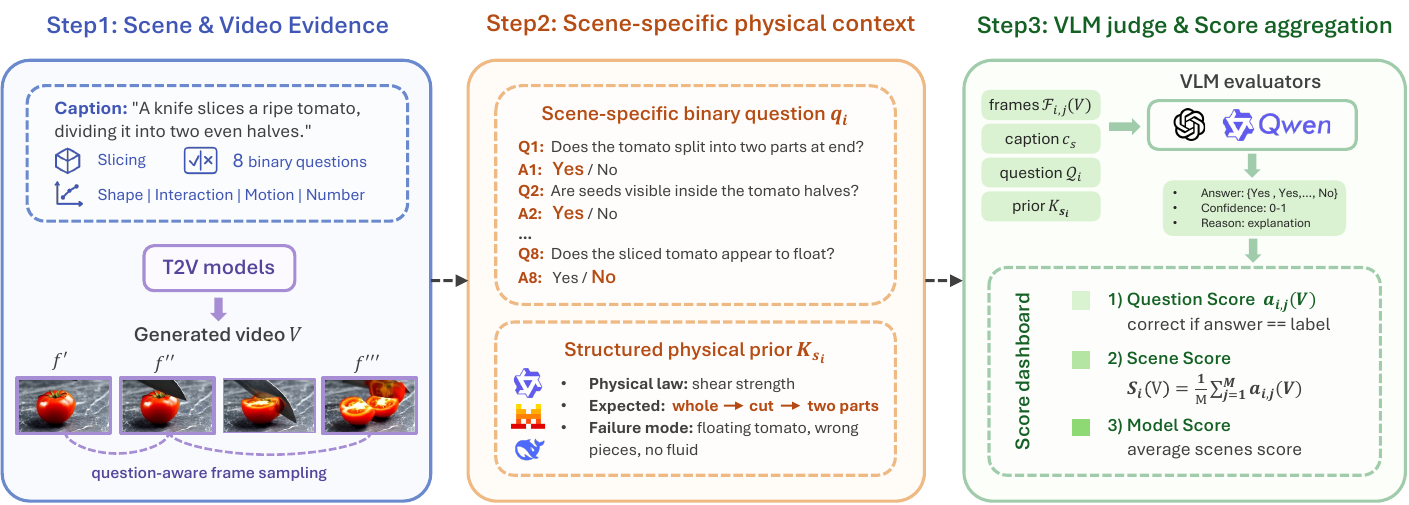}
    \caption{\small\textbf{\phymetric{} evaluation pipeline.}
    Given a physical scene $s$, a text-to-video model generates a video $V_s$ from the scene caption $c_s$.
    For each scene-specific binary question, \phymetric{} samples question-aware frames and provides a structured physical prior $K_s$ describing the expected law, outcome, and common failure modes.
    A VLM then answers each question conditioned on the sampled frames and $K_s$.
    Question correctness is averaged into the scene score $S(s,V_s)$, and model performance is obtained by averaging scene scores over all benchmark scenes.
    This pipeline turns holistic video evaluation into grounded, interpretable physical diagnosis.
    }
    \label{fig:evaluation_pipeline}
    \vspace{-10pt}
\end{figure}

\subsection{\phymetric{} Main Comparison}
\vspace{-0.2cm}
\label{subsec:phymetricResults}

\paragraph{Full benchmark results.}
Table~\ref{tab:results_full} reports results on the full benchmark using two open-source VLM judges, Qwen3.5-9B~\cite{yang2025qwen3} and Qwen2.5-7B~\cite{bai2025qwen2}.
\phymetric{} separates current T2V models under both judges.
Wan2.2-14B obtains the best average score, followed closely by Cosmos-7B.
Wan2.1-1.3B ranks third, while CogVideoX-5B obtains the lowest score.
The ranking is not explained by parameter count alone.
CogVideoX-5B is larger than Wan2.1-1.3B, but scores lower under both judges.
This suggests that \phymetric{} captures model differences in physical behaviors, rather than only model scale.
The gap between the best and weakest models also shows that current open-source T2V models are not saturated on \phymetric{}.

\paragraph{Controlled subset results.}
Table~\ref{tab:results_subset} reports results on the 100-scene subset with three VLM judges, Qwen3.5-9B~\cite{yang2025qwen3}, Qwen2.5-7B~\cite{bai2025qwen2}, and GPT-5.4~\cite{singh2025openai_gpt5}.
The ranking is broadly consistent with the full benchmark.
Wan2.2-14B and Cosmos-7B form the strongest group, while CogVideoX-5B remains the weakest model.
The absolute scores differ across judges.
Qwen2.5-7B is more conservative than Qwen3.5-9B and GPT-5.4, but the model ordering remains stable enough to support cross-evaluator comparison.
The controlled subset also shows why a single score is not sufficient.
Cosmos-7B slightly exceeds Wan2.2-14B in the three-judge average, while Wan2.2-14B is stronger under Qwen3.5-9B, that suggests close overall scores may hide different physical capability profiles.

\subsection{Dimension-Level Analysis}
\begin{figure*}
    \centering
    \includegraphics[width=1\linewidth]{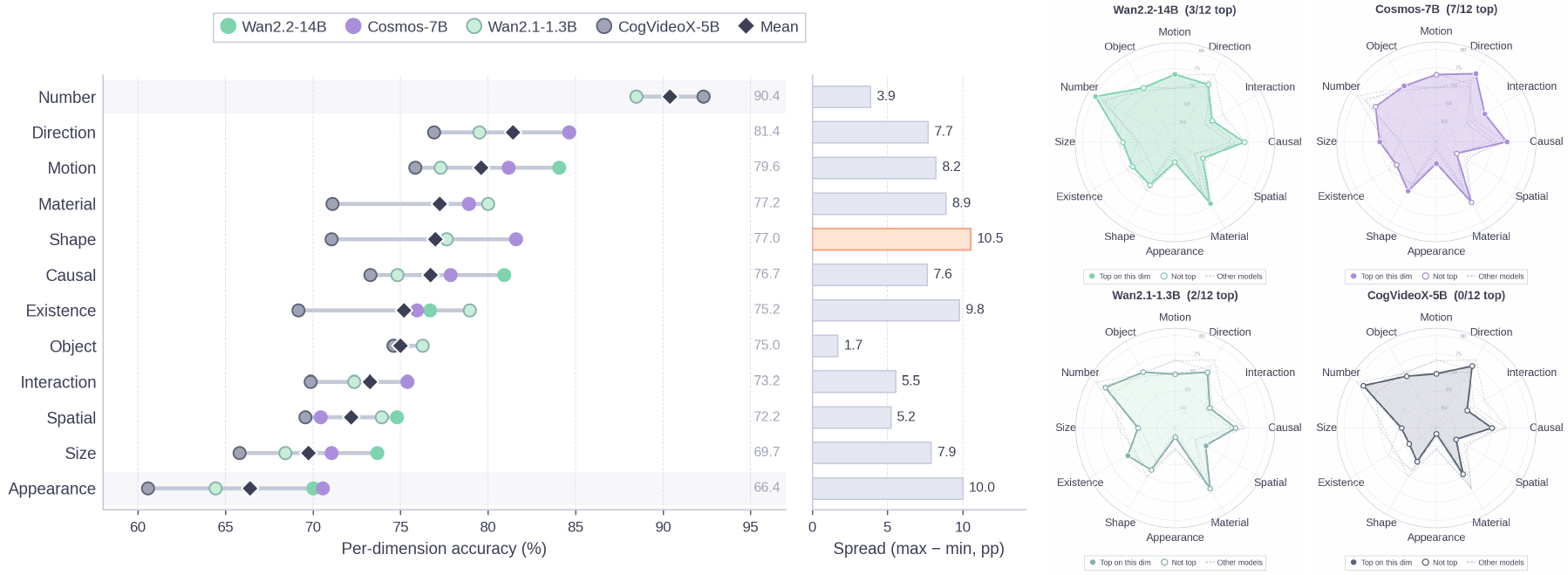}
    \caption{\small \textbf{Per-dimension physical performance of current T2V models.}
    Accuracy across 12 evaluation dimensions for four T2V models, with diamonds denoting mean accuracy.
    Radar plots showing each model's dimension profile and the dimensions where it achieves the best performance.}
    \label{fig:dimensions_analysis}
    \vspace{-10pt}
\end{figure*}

\vspace{-0.2cm}
\label{subsec:dimensionAnalysis}
Figure~\ref{fig:dimensions_analysis} reveals that current T2V models exhibit highly uneven physical capabilities across the 12 dimensions, with average accuracy ranging from 90.39\% on \textit{number} down to 66.39\% on \textit{appearance}.
The strongest performance appears in \textit{number}, suggesting that current T2V models and VLM judges can reliably verify object count when visual evidence is explicit.
\textit{Direction} and \textit{motion} also remain relatively robust, both exceeding 79\%, which indicates that coarse kinematic patterns and large-scale movement trends are generally captured more consistently than deeper physical attributes.
In contrast, the lowest accuracy on \textit{appearance} reveals that physical failures frequently emerge through material, state, or surface inconsistency rather than purely incorrect motion trajectories.
Similarly, weaker performance on \textit{size} and \textit{spatial} dimensions indicates persistent difficulty in preserving object scale and relative placement over time.

Inter-model spread further highlights which dimensions most effectively separate current generators.
\textit{Appearance} and \textit{shape} show the largest spreads (10.00 and 10.53 points), suggesting that material realism and deformation-sensitive physical behavior remain major differentiators across models.
\textit{object} identity exhibits only a 1.69-point spread, implying that current generators are comparatively similar on basic object-presence checks while diverging more substantially on physically meaningful appearance and structural consistency.

We further visualize the dimension profiles of the four evaluated T2V models.
Wan2.2-14B forms the broadest and most stable profile, with strong coverage on kinematic and causal dimensions.
Cosmos-7B attains the largest number of dimension-wise best scores, indicating that it is highly competitive on specific physical checks.
Wan2.1-1.3B shows a more selective profile on material and existence dimensions,
while CogVideoX-5B shows a visibly contracted profile, especially in dimensions that require stable physical appearance or relational evidence.
This visualization suggests that model differences are not only scalar, but also structural, in which two models with close overall scores may succeed on different types of physical evidence.

\begin{table*}[t]
\centering
\begin{minipage}{0.43\textwidth}
    \centering
    \caption{Main results on the full \phymetric{} benchmark (1,115 scenes) using open-source VLM judges. \emph{Average} denotes the mean performance across VLM judges.}
    \label{tab:results_full}
    \small
    \setlength{\tabcolsep}{2pt}
    \renewcommand{\arraystretch}{1.05}
    \resizebox{\textwidth}{!}{
    \begin{tabular}{lcccccc}
    \toprule
    \multirow{2}{*}{\textbf{Model}} & \multicolumn{2}{c}{\textbf{Qwen3.5-9B}} & \multicolumn{2}{c}{\textbf{Qwen2.5-VL-7B}} & \multicolumn{2}{c}{\textbf{Average}} \\
    \cmidrule(lr){2-3} \cmidrule(lr){4-5} \cmidrule(lr){6-7}
    & Overall & Weighted & Overall & Weighted & Overall & Weighted \\
    \midrule
    Wan2.2-14B   & \textbf{76.74} & \textbf{78.07} & 68.82          & 70.82          & \textbf{72.78} & \textbf{74.45}  \\
    Cosmos-7B    & 76.28          & 77.60          & \textbf{68.89} & \textbf{70.92} & 72.59          & 74.26  \\
    Wan2.1-1.3B  & 73.86          & 75.50          & 67.26          & 69.42          & 70.56          & 72.46  \\
    CogVideoX-5B & 71.36          & 73.18          & 65.12          & 67.62          & 68.24          & 70.40  \\
    \bottomrule
    \end{tabular}
    }
\end{minipage}
\hfill 
\begin{minipage}{0.54\textwidth}
    \centering
    \caption{Evaluation on the 100-scene controlled subset across both closed and open-source VLM evaluators. \emph{Average} reflects the mean accuracy across all three judges.}
    \label{tab:results_subset}
    \small
    \setlength{\tabcolsep}{2pt}
    \renewcommand{\arraystretch}{1.05}
    \resizebox{\textwidth}{!}{%
    \begin{tabular}{lcccccccc}
    \toprule
    \multirow{2}{*}{\textbf{Model}} & \multicolumn{2}{c}{\textbf{Qwen3.5-9B}} & \multicolumn{2}{c}{\textbf{Qwen2.5-VL-7B}} & \multicolumn{2}{c}{\textbf{GPT-5.4}} & \multicolumn{2}{c}{\textbf{Average}} \\
    \cmidrule(lr){2-3} \cmidrule(lr){4-5} \cmidrule(lr){6-7} \cmidrule(lr){8-9}
    & Overall & Weighted & Overall & Weighted & Overall & Weighted & Overall & Weighted \\
    \midrule
    Wan2.2-14B   & \textbf{77.88} & \textbf{79.23} & 66.75          & 69.54          & 73.12          & 74.26          & 72.58          & 74.34  \\
    Cosmos-7B    & 76.50          & 77.54          & \textbf{67.37} & \textbf{70.37} & \textbf{74.00} & \textbf{75.54} & \textbf{72.62} & \textbf{74.48}  \\
    Wan2.1-1.3B  & 74.62          & 76.85          & 65.87          & 68.94          & 72.12          & 73.88          & 70.87          & 73.22  \\
    CogVideoX-5B & 70.63          & 72.45          & 64.38          & 67.13          & 69.75          & 71.67          & 68.25          & 70.42  \\
    \bottomrule
    \end{tabular}%
    }
\end{minipage}
\end{table*}

\subsection{Ablation Study}
\vspace{-0.2cm}
\label{subsec:ablation}

We conduct ablations on the 100-scene subset to examine the two main design choices of \phymetric{}: scene-specific QA and the structured physical prior.
For each scene $s$, the structured physical prior $K_s$ is generated before video evaluation from the scene description and category metadata.
Importantly, $K_s$ is generated without access to the generated videos or ground-truth question answers, which provides scene-level expectation grounding VLM judge.

\paragraph{Component ablation.}

Table~\ref{tab:ablation_components} evaluates two components of \phymetric{}, the scene-specific question design and the structured physical priors $K_s$.
The results show that scene-specific QA is the key component for reliable physical diagnosis.
When scene-specific questions are replaced by the generic query \textit{``Is this video physically plausible?''}, accuracy drops to near chance for all evaluators.
Without $K_s$, Qwen3.5-9B drops from 72.75\% to 48.13\%, Qwen2.5-VL from 63.47\% to 47.31\%, and GPT-5.4 from 73.94\% to 49.25\%.
Adding $K_s$ to the generic query still fails to recover the missing signal, indicating that physical context alone is insufficient when the evaluator is not asked to verify a concrete scene-specific consequence.

When scene-specific questions are kept, $K_s$ provides a complementary but evaluator-dependent benefit.
For Qwen3.5-9B, adding $K_s$ improves both accuracy and human correlation, increasing accuracy from 72.75\% to 74.91\% and $\rho_{\text{human}}$ from 0.165 to 0.201.
For Qwen2.5-VL, $K_s$ improves accuracy from 63.47\% to 66.09\%, though human correlation decreases.
For GPT-5.4, $K_s$ slightly lowers label accuracy but improves human correlation from 0.216 to 0.250.
These results suggest that benchmark-label accuracy and human alignment measure related but distinct evaluator behaviors.
Overall, scene-specific QA is necessary for diagnostic accuracy, while $K_s$ can further improve alignment when the evaluator can use the prior effectively.



\begin{table*}[t]
  \centering
  \begin{minipage}{0.48\textwidth}
    \centering
    \caption{\small\textbf{Component ablation of \textsc{PhyMetric}.} 
    We toggle the structured physical prior $K_s$ and the question style. 
    ``Scene-specific'' denotes targeted binary questions, while ``Generic'' 
    replaces each with a single global query.}
    \label{tab:ablation_components}
    \small
    \setlength{\tabcolsep}{4pt} 
    \renewcommand{\arraystretch}{1.32}
    \resizebox{\textwidth}{!}{
    \begin{tabular}{ccl ccc} 
      \toprule
      \textbf{$K_s$} & \textbf{Question} & \textbf{Evaluator}
      & \textbf{Overall} & \textbf{Weighted}
      & \textbf{$\rho_{\text{human}}$} \\ 
      \midrule
      \ding{51} & Scene-specific & Qwen3.5-9B & \textbf{74.91} & \textbf{76.52} & \textbf{+0.201} \\
      \ding{55} & Scene-specific & Qwen3.5-9B & 72.75 & 74.56 & +0.165 \\
      \ding{55} & Generic        & Qwen3.5-9B & 48.13 & 49.71 & +0.192 \\
      \ding{51} & Generic        & Qwen3.5-9B & 48.88 & 50.17 & +0.196 \\
      \midrule
      \ding{51} & Scene-specific & Qwen2.5-VL & \textbf{66.09} & \textbf{69.00} & +0.104 \\
      \ding{55} & Scene-specific & Qwen2.5-VL & 63.47 & 65.89 & \textbf{+0.215} \\
      \ding{55} & Generic        & Qwen2.5-VL & 47.31 & 50.36 & +0.139 \\
      \ding{51} & Generic        & Qwen2.5-VL & 46.91 & 50.64 & +0.030 \\
      \midrule
      \ding{51} & Scene-specific & GPT-5.4 & 72.25 & 73.84 & \textbf{+0.250}$^{*}$ \\
      \ding{55} & Scene-specific & GPT-5.4 & \textbf{73.94} & \textbf{75.28} & +0.216 \\
      \ding{55} & Generic        & GPT-5.4 & 49.25 & 49.30 & +0.121 \\
      \ding{51} & Generic        & GPT-5.4 & 47.12 & 48.45 & +0.202 \\
      \bottomrule
    \end{tabular}}
  \end{minipage}
  \hfill
  \begin{minipage}{0.48\textwidth}
    \centering
    \caption{\small\textbf{Effect of structured physical prior source and evaluator.} 
    Each row is a pair of prior source and question evaluator. 
    \emph{zero-shot} means without the structured physical prior $K_s$.}
    \label{tab:ablation_kb_eval}
    \small
    \setlength{\tabcolsep}{4pt}
    \renewcommand{\arraystretch}{1.05}
    \resizebox{\textwidth}{!}{
    \begin{tabular}{ll ccc}
      \toprule
      \textbf{$K_s$ source} & \textbf{Evaluator}
      & \textbf{Overall} & \textbf{Weighted}
      & \textbf{$\rho_{\text{human}}$} \\ 
      \midrule
      \textit{zero-shot} & Qwen3.5-9B & 72.75 & 74.56 & +0.165 \\
      Qwen3.5-9B & Qwen3.5-9B & \textbf{74.91} & \textbf{76.52} & +0.201 \\
      Qwen2.5-7B & Qwen3.5-9B & 73.66 & 75.42 & +0.209 \\
      DeepSeek-R1 & Qwen3.5-9B & 73.81 & 75.67 & \textbf{+0.251}$^{*}$ \\
      Mistral-7B & Qwen3.5-9B & 74.09 & 76.13 & +0.248$^{*}$ \\
      \midrule
      \textit{zero-shot} & Qwen2.5-VL & 63.47 & 65.89 & +0.015 \\
      Qwen3.5-9B & Qwen2.5-VL & \textbf{66.09} & \textbf{69.00} & +0.104 \\
      Qwen2.5-7B & Qwen2.5-VL & 63.25 & 66.15 & +0.156 \\
      DeepSeek-R1 & Qwen2.5-VL & 61.97 & 64.53 & \textbf{+0.194} \\
      Mistral-7B & Qwen2.5-VL & 62.91 & 65.50 & +0.158 \\
      \midrule
      \textit{zero-shot} & GPT-5.4 & \textbf{73.94} & \textbf{75.28} & +0.216 \\
      Qwen3.5-9B & GPT-5.4 & 72.25 & 73.84 & +0.250$^{*}$ \\
      Qwen2.5-7B & GPT-5.4 & 72.72 & 73.87 & +0.283$^{*}$ \\
      DeepSeek-R1 & GPT-5.4 & 72.47 & 73.76 & +0.270$^{*}$ \\
      Mistral-7B & GPT-5.4 & 72.56 & 73.64 & \textbf{+0.293}$^{*}$ \\
      \bottomrule
    \end{tabular}}
  \end{minipage}
  \vspace{0pt}
  \begin{flushleft}
    \hspace{0.02\textwidth} \footnotesize{$^{*}$ marks $p<0.05$.}
  \end{flushleft}
  \vspace{-15pt}
\end{table*}

\paragraph{Effect of structured physical prior source and evaluator.}
To test whether the structured physical prior depends on a particular LLM source, we examine $K_s$ with different auxiliary LLMs and evaluate them with different VLM judges in Table~\ref{tab:ablation_kb_eval}.
Across evaluators, the best human correlation often comes from a prior source different from the evaluator itself.
For Qwen3.5-9B, all structured priors improve over zero-shot, and DeepSeek-R1 and Mistral-7B achieve the highest human correlations, with $\rho_{\text{human}}=0.251$ and $0.248$.
For Qwen2.5-VL, DeepSeek-R1 also gives the strongest human correlation.
For GPT-5.4, zero-shot gives the highest raw accuracy, but every structured prior improves human correlation, with Mistral-7B reaching the best $\rho_{\text{human}}=0.293$.
This cross-source pattern suggests that $K_s$ is not simply a self-consistency boost from the same model family.
Instead, structured physical priors provide an external expectation signal that can make VLM judgments more aligned with human-perceived physical plausibility.

\noindent\textbf{Findings}
From the main results on full benchmark and subset, \phymetric{} separates current T2V models and reveals that better visual quality does not always imply stronger physical behavior.
Second, the dimension scores show that current models are stronger at coarse motion and object count, but still struggle with higher-order physical dimensions, including material state, appearance stability, scale consistency, and spatial relations.
Third, the ablation results confirm that scene-specific QA is necessary for meaningful physical evaluation, while structured physical priors provide a useful but evaluator-dependent grounding signal.

%% file: sections/07conclusion.tex
\section{Conclusion}
\vspace{-0.2cm}
\label{sec:con}

We present \phymetric{}, a structured benchmark for evaluating physical plausibility in T2V generation through scene-grounded diagnostic question answering.
Unlike prior evaluation paradigms that rely primarily on holistic plausibility or coarse realism judgments, \phymetric{} decomposes physical understanding into verifiable scene-specific consequences across diverse physical domains, dimensions, and difficulty levels.
By combining a large-scale taxonomy of 50 physical categories, expert-filtered benchmark construction, structured physical priors, question-aware frame sampling, and hierarchical score aggregation, \phymetric{} provides both broad coverage and fine-grained diagnostic interpretability.
Our experiments show that physical evaluation is fundamentally more reliable when grounded in targeted physical consequences rather than generic plausibility.
Across multiple VLM judges, \phymetric{} consistently reveals stable model rankings and systematic weaknesses in current T2V systems.
These findings suggest that our \phymetric{} can serve not only as a benchmark for model comparison, but also as a diagnostic framework for guiding future progress in physically grounded video generation.

\begin{ack}
This work was supported in part by the National Science Foundation under Award No.~2511989 for the MRI project ``Acquisition of a High-Performance GPU Cluster for Large-Scale AI''.
Any opinions, findings, and conclusions or recommendations expressed in this material are those of the authors and do not necessarily reflect the views of the funding agencies.
The authors declare no competing interests.
\end{ack}

%% file: checklist.tex
\section*{NeurIPS Paper Checklist}

\begin{enumerate}

\item {\bf Claims}
    \item[] Question: Do the main claims made in the abstract and introduction accurately reflect the paper's contributions and scope?
    \item[] Answer: \answerYes{} 
    \item[] Justification: We describe our benchmark's contribution in abstract and introduction section in detail. Our PhyMetric provides a diagnostic benchmark for video physical plausibility evaluation.

\item {\bf Limitations}
    \item[] Question: Does the paper discuss the limitations of the work performed by the authors?
    \item[] Answer: \answerYes{} 
    \item[] Justification: We have claimed our limitation in \autoref{appen:limitation}.

\item {\bf Theory assumptions and proofs}
    \item[] Question: For each theoretical result, does the paper provide the full set of assumptions and a complete (and correct) proof?
    \item[] Answer: \answerNA{} 
    \item[] Justification: The paper does not include theoretical results.

    \item {\bf Experimental result reproducibility}
    \item[] Question: Does the paper fully disclose all the information needed to reproduce the main experimental results of the paper to the extent that it affects the main claims and/or conclusions of the paper (regardless of whether the code and data are provided or not)?
    \item[] Answer: \answerYes{} 
    \item[] Justification: The experimental implementation details are described in \autoref{sec:exp}, and all datasets used in this work are open accessed online and they are referenced soundly in paper.

\item {\bf Open access to data and code}
    \item[] Question: Does the paper provide open access to the data and code, with sufficient instructions to faithfully reproduce the main experimental results, as described in supplemental material?
    \item[] Answer: \answerYes{} 
    \item[] Justification: We have released the evaluation scripts in the anonymous code site \hyperlink{https://anonymous.4open.science/r/PhyMetric-Benchmark/}{https://anonymous.4open.science/r/PhyMetric-Benchmark/}. We also provide sample 10 scenes across four video generators, three VLM judges, and human evaluation in the supplementary material.

\item {\bf Experimental setting/details}
    \item[] Question: Does the paper specify all the training and test details (e.g., data splits, hyperparameters, how they were chosen, type of optimizer) necessary to understand the results?
    \item[] Answer: \answerYes{} 
    \item[] Justification: We provide detailed dataset usage in \autoref{sec:exp}.

\item {\bf Experiment statistical significance}
    \item[] Question: Does the paper report error bars suitably and correctly defined or other appropriate information about the statistical significance of the experiments?
    \item[] Answer: \answerNo{} 
    \item[] Justification: Our evaluation uses protocols from previous benchmarks are not accompanied by error bars, confidence intervals, nor statistical significance tests.

\item {\bf Experiments compute resources}
    \item[] Question: For each experiment, does the paper provide sufficient information on the computer resources (type of compute workers, memory, time of execution) needed to reproduce the experiments?
    \item[] Answer: \answerYes{} 
    \item[] Justification: We provide the detailed computation resources to reproduce the experiments in \autoref{sec:exp} and the evaluation code on three VLM judges is provided in the anonymous code. Please refer to Q5.
    
\item {\bf Code of ethics}
    \item[] Question: Does the research conducted in the paper conform, in every respect, with the NeurIPS Code of Ethics \url{https://neurips.cc/public/EthicsGuidelines}?
    \item[] Answer: \answerYes{} 
    \item[] Justification: The research conducted in the paper conforms with the NeurIPS Code of Ethics.

\item {\bf Broader impacts}
    \item[] Question: Does the paper discuss both potential positive societal impacts and negative societal impacts of the work performed?
    \item[] Answer: \answerYes{} 
    \item[] Justification: This paper provides a scene-level diagnostic benchmark for evaluating physical plausibility in text-to-video generation.
    
\item {\bf Safeguards}
    \item[] Question: Does the paper describe safeguards that have been put in place for responsible release of data or models that have a high risk for misuse (e.g., pre-trained language models, image generators, or scraped datasets)?
    \item[] Answer: \answerNA{} 
    \item[] Justification: This work poses no such risks.

\item {\bf Licenses for existing assets}
    \item[] Question: Are the creators or original owners of assets (e.g., code, data, models), used in the paper, properly credited and are the license and terms of use explicitly mentioned and properly respected?
    \item[] Answer: \answerYes{} 
    \item[] Justification: We cited all works of codes and datasets used in the paper.

\item {\bf New assets}
    \item[] Question: Are new assets introduced in the paper well documented and is the documentation provided alongside the assets?
    \item[] Answer: \answerYes{} 
    \item[] Justification: \phymetric{} introduces a new benchmark dataset with scene definitions, QA annotations, physical priors, and evaluation code, documented in \autoref{sec:phymetric} and Appendix~\ref{appen:taxonomy}.

\item {\bf Crowdsourcing and research with human subjects}
    \item[] Question: For crowdsourcing experiments and research with human subjects, does the paper include the full text of instructions given to participants and screenshots, if applicable, as well as details about compensation (if any)? 
    \item[] Answer: \answerYes{} 
    \item[] Justification: The paper includes a human validation study in which annotators answer the same scene-specific yes/no questions used by \phymetric{} on the controlled 100-scene subset. We provide the annotation instructions, interface screenshots, quality-control protocol, and compensation details in Appendix~\ref{appen:human_study}.

\item {\bf Institutional review board (IRB) approvals or equivalent for research with human subjects}
    \item[] Question: Does the paper describe potential risks incurred by study participants, whether such risks were disclosed to the subjects, and whether Institutional Review Board (IRB) approvals (or an equivalent approval/review based on the requirements of your country or institution) were obtained?
    \item[] Answer: \answerNo{} 
    \item[] Justification: This study involved low-risk benchmarking annotation only and did not require formal IRB approval under local institutional guidance.

\item {\bf Declaration of LLM usage}
    \item[] Question: Does the paper describe the usage of LLMs if it is an important, original, or non-standard component of the core methods in this research? Note that if the LLM is used only for writing, editing, or formatting purposes and does \emph{not} impact the core methodology, scientific rigor, or originality of the research, declaration is not required.
    \item[] Answer: \answerYes{} 
    \item[] Justification: We employ VLMs, including GPT-5.4, Qwen3.5-9B, Qwen2.5-7B, and LLMs, including DeepSeek-R1-8B, and Mistral-7B as the reasoning tool for our benchmark, as mentioned in \autoref{sec:exp}. The usage of LLMs and VLMs only for experimental reasoning.

\end{enumerate}